\pdfoutput=1

\documentclass[11pt]{article}

\usepackage{booktabs}
\usepackage{multirow}
\usepackage{graphicx}
\usepackage[table,xcdraw]{xcolor}
\usepackage{ulem}
\usepackage{subcaption}
\usepackage{kotex}

\usepackage{pifont}
\usepackage{soul}
\usepackage[preprint]{acl}

\usepackage{times}
\usepackage{latexsym}

\usepackage[T1]{fontenc}

\usepackage{amsmath}

\usepackage{amssymb}

\usepackage[utf8]{inputenc}

\usepackage{microtype}

\usepackage{inconsolata}

\usepackage{graphicx}

\usepackage{xspace}

\newcommand{\pa}{Pattern-Aware\xspace}
\newcommand{\pld}{PLD\xspace}
\newcommand{\ptf}{PTF\xspace}

\usepackage{comment}

\title{Beyond Line-Level Filtering for the Pretraining Corpora of LLMs}

\author{
  Chanwoo Park\thanks{Dept. of Computer Science, Seoul National University} \\
  \texttt{99chanwoo@snu.ac.kr} \\\And
  Suyoung Park\thanks{Graduate School of Data Science, Seoul National University} \\
  \texttt{itforspark@snu.ac.kr} \\\And
  Yelim Ahn\footnotemark[2] \\
  \texttt{mileya@snu.ac.kr} \\\AND
  Jongmin Kim\footnotemark[2] \\
  \texttt{jmin.kim@snu.ac.kr} \\\And
  Jongyeon Park\footnotemark[2] \\
  \texttt{iscopark67@snu.ac.kr} \\\And
  Jaejin Lee\footnotemark[1]\footnotemark[2] \\
  \texttt{jaejin@snu.ac.kr}
  }

\newcommand{\base}[0]{\textbf{base(1B)}}
\newcommand{\sm}[0]{\textbf{small(900M)}}

\newcommand{\tablecell}[1]{\begin{tabular}[t]{@{}p{\linewidth}@{}}#1\end{tabular}}

\begin{document}
\maketitle

\begin{abstract}
While traditional line-level filtering techniques, such as line-level deduplication and trailing-punctuation filters, are commonly used, these basic methods can sometimes discard valuable content, negatively affecting downstream performance. In this paper, we introduce two methods—pattern-aware line-level deduplication (PLD) and pattern-aware trailing punctuation filtering (PTF)—by enhancing the conventional filtering techniques. Our approach not only considers line-level signals but also takes into account their sequential distribution across documents, enabling us to retain structurally important content that might otherwise be removed. We evaluate these proposed methods by training small language models (1 B parameters) in both English and Korean. The results demonstrate that our methods consistently improve performance on multiple-choice benchmarks and significantly enhance generative question-answering accuracy on both SQuAD v1 and KorQuAD v1.\footnote{Our code can be found at https://github.com/mcrl/pattern-aware-filtering}

\end{abstract}
\section{Introduction}
\label{sec:introduction}
Large Language Models (LLMs) have shown impressive performance across various natural language processing (NLP) tasks and real-world applications. A significant breakthrough has been achieved with GPT~\cite{gpt2,oaigpt3}, which emphasizes the importance of increasing model size. However, subsequent research revealed that proportional increases must match this scaling in training data~\cite{hoffmann2022training, muennighoff2023scaling}. 

Following this insight, recent foundation models such as LLaMA~\cite{touvron2023llama}, Qwen~\cite{qwen2025qwen25technicalreport}, and Gemma~\cite{gemmateam2025gemma3technicalreport} have been trained on trillions of tokens, although the specifics of their training corpora remain unpublished. In response to this gap, several initiatives have sought to open-source large-scale corpora~\cite{oscar2022, penedo2023refinedweb, penedo2024fineweb, soldaini2024dolma}, providing valuable resources for research on LLM pretraining~\cite{almazrouei2023falconseriesopenlanguage}, model architectures~\cite{gu2023mamba}, training methodologies~\cite{kim2023solar}, and data deduplication~\cite{lee2021deduplicating}.

Web data, with its vastness and variety, has become essential for constructing open-source datasets. CommonCrawl~\footnote{https://commoncrawl.org/} is one of the most widely used sources for training LLMs~\cite{gao2020pile, penedo2023refinedweb, raffel2023exploring, penedo2024fineweb, soldaini2024dolma, li2024datacomp}. Its WET (WARC Encapsulated Text) distribution, which consists of plain text extracted from HTML, has been the foundational resource for pioneering models~\cite{oaigpt3, raffel2023exploring, touvron2023llama} and remains valuable for organizations with limited resources. Although this distribution provides over 250 B documents in less than 1 PB of disk space, many lines are irrelevant and unlikely to enhance the downstream performance of the models. To mitigate this issue, simple filtering techniques—such as line-level deduplication and trailing punctuation filters—are commonly applied as preprocessing steps to extract more useful text.

While heuristic rules can provide valuable signals regarding the importance of certain lines, relying solely on these signals to determine whether a line should be removed has clear limitations. For example, assuming that lines repeated across documents should be discarded, or that lines lacking trailing punctuation are uninformative, can result in the loss of genuinely useful content. Our analysis of the raw data indicates that these rules frequently eliminate lines that could enhance downstream performance. This underscores the necessity of considering not just line-level signals, but also their distribution throughout the document.

To this end, this paper proposes a method called \textit{pattern-aware line filtering} for extracting relevant text from the CommonCrawl WET dataset. Our approach is based on two rules: \textit{pattern-aware line-level deduplication (\pld)} and \textit{pattern-aware trailing-punctuation filtering (\ptf)}.  \pld categorizes each line according to the number of documents in which it appears, allowing us to distinguish between distinctive lines and repetitive or boilerplate content. Meanwhile, \ptf classifies lines based on the presence of trailing punctuation marks, helping to identify sentence-like structures. Rather than filtering lines in isolation, \pld and \ptf view an entire document as a sequence of categorization results, which helps detect structural patterns indicative of relevant text. This holistic approach allows us to retain important information that might otherwise be discarded by traditional line-wise filtering methods.



We evaluate our method by training small LLMs and demonstrate that pattern-aware line filtering enhances performance in downstream tasks, including generative QA benchmarks, such as SQuAD v1~\cite{rajpurkar2016squad1} and KorQuAD v1~\cite{lim2019korquad1}. Furthermore, we show its effectiveness in both English and Korean, indicating its applicability to multilingual corpora. The contributions of this paper are summarized as follows:
\begin{itemize}
\setlength{\itemsep}{0pt}
\setlength{\parskip}{0pt}     
\setlength{\parsep}{0pt}
    \item We revisit heuristic filtering rules and highlight cases where lines dismissed as low-quality actually contain valuable content that can enhance the capabilities of LLMs.
    \item We introduce a new method called \textit{pattern-aware line filtering}, which uses both line-level signals and document-level distributions to more effectively retain relevant text.
    \item We evaluate our method on English and Korean corpora, demonstrating the model's improved performance in downstream tasks and showing that our approach generalizes across different languages.
    \item We provide a fully reproducible pipeline for dataset construction. The source data is publicly available, and we will also make our code publicly accessible. To our knowledge, this is the first fully open and reproducible filtering technique specifically designed for large-scale processing of Korean CommonCrawl WET data.
\end{itemize}

\section{Pretraining Datasets for LLMs}
\label{sec:related-work}

\paragraph{Large-scale pretraining text datasets.}
State-of-the-art LLMs are frequently pretrained on trillions of tokens sourced from proprietary datasets. However, several initiatives have released open-source large-scale corpora, which provide valuable resources for the research community~\cite{penedo2023refinedweb,raffel2023exploring,soldaini2024dolma,penedo2024fineweb,li2024datacomp,weber2024redpajama}. Most of these corpora are derived from the petabyte-scale CommonCrawl web archive, which contains more than 250 B webpages collected since 2008. The openness and reproducibility of this data have enabled researchers to share not only the raw data but also the preprocessing pipelines, promoting iterative improvements across various projects. Among the various distributions of CommonCrawl, the WET format is the most practical choice for groups with limited resources. By providing plain text extracted from HTML, it reduces storage requirements by six times compared to the original WARC files. This cost efficiency enables research groups to access hundreds of billions of documents and has made WET the backbone of many large-scale open corpora, including C4~\cite{raffel2023exploring}, Dolma ~\cite{soldaini2024dolma}, Yi~\cite{young2024yi}, and RedPajama~\cite{weber2024redpajama}.

\paragraph{Korean corpora for pretraining LLMs.}
As the development of large language models (LLMs) continues globally, South Korean companies have also been actively creating their own models. Notable examples include Naver’s HyperCLOVA \cite{yoo2024hyperclova}, Kakao’s Kanana \cite{kananallmteam2025kanana}, and LG AI Research’s Exaone \cite{exaone-4.0}. Similar to their international counterparts, these models are trained on proprietary corpora, which often lack transparency regarding the composition and preprocessing methods of the data used. 

While multilingual corpora such as mC4~\cite{xue2020mt5}, OSCAR~\cite{oscar2022}, and FineWeb~\cite{penedo2024fineweb} contain Korean subsets and can serve as open alternatives, their processing pipelines are largely optimized for English, frequently overlooking the specific characteristics of other languages. Our analysis indicates that filtering steps can yield significantly different effects across languages. To address this, we have tailored our filtering methodology to suit both English and Korean separately, ensuring that each language benefits from customized preprocessing.

Open-source corpus construction projects often use data ablation, which involves training models with the same setup but different preprocessing steps. This allows researchers to isolate the effects of each refinement, an approach that is essential for making informed decisions about pipeline design. However, proprietary data pipelines for Korean LLMs~\cite{SKTAdotX3.1, exaone-4.0, cha2025varcovision20technicalreport} typically obscure their filtering methods and the rationale behind their designs. To our knowledge, Kanana~\cite{kananallmteam2025kanana} is the only Korean project that reports on data ablation for preprocessing pipeline decisions. Unfortunately, their results are not reproducible due to the use of proprietary raw data.

In contrast, our work aims to offer a fully reproducible method for constructing Korean corpora, one that can also be adapted to multilingual settings. This will enable future researchers to build upon, evaluate, and extend our approach.
\section{Observations based on Data Ablation}
\label{sec:methodology}
To make informed decisions about filtering rules, we conduct data ablation studies. In these studies, language models are trained on datasets that have been filtered using the target technique and then evaluated on various downstream tasks. By comparing evaluation results from different filtering choices with baseline results, while keeping the model architecture and training recipe consistent, we can analyze how different line filtering techniques affect downstream performance. To assess their impact in both English and Korean, we have established separate experimental setups for each language.

\subsection{Training Datasets \label{sec:training-data}}
We use CommonCrawl data snapshots, with each snapshot representing one month of data, from 2019 to 2022. For each year, we select the first five snapshots, resulting in a total of 20 snapshots. To construct the baseline dataset for each language, we apply language identification and retain only documents in the target language. An English document is classified as such if it has a classification score of at least 0.65 from the FastText~\cite{joulin2016fasttext} language classifier. A Korean document is defined as one in which at least 10\% of the characters are Hangul. Additional details regarding the training dataset can be found in Appendix~\ref{sec:app-detail-data}.

\subsection{Model Training \label{sec:training-data}}
We train decoder-only transformer models based on the architecture of LLaMA 3~\cite{grattafiori2024llama3herdmodels}. For our English models, we use the LLaMA 3 tokenizer, while for the Korean models, we employ the Exaone 4.0 tokenizer~\cite{exaone-4.0}. This tokenizer is widely adopted in research on Korean-capable LLMs and has a vocabulary size comparable to that of LLaMA 3. Our models range from 1.3 B to 1.5 B parameters, which we refer to as the \textbf{base(1B)} model. This selection is in line with the model sizes typically used in previous ablation studies~\cite{soldaini2024dolma}. To assist with threshold selection in our pattern-aware filtering rules, we also train smaller models that range from 880 M to 960 M parameters, referred to as the \textbf{small(900M)} model. Across both scales, we maintain a consistent number of training tokens by adhering to the Chinchilla-optimal data ratio~\cite{hoffmann2022training}: we use 30 billion tokens for the \base ~and 20 billion tokens for \sm. Further training details can be found in Appendix~\ref{sec:app-detail-model}.

\subsection{Downstream Tasks\label{sec:training-data}}
For selecting downstream tasks, we prioritize benchmarks that: (i) evaluate model capabilities at \sm ~and \base ~scales, especially for generative tasks; (ii) encompass a diverse range of language model knowledge; and (iii) have been widely used in previous evaluations of language model pretraining, regardless of the language.

Based on these criteria, we select eight downstream tasks for English and five for Korean.  
In English, we use SQuAD v1~\cite{rajpurkar2016squad1}, Hellaswag~\cite{zellers2019hellaswag}, ARC-Easy~\cite{arc}, PIQA~\cite{bisk2020piqa}, SciQ~\cite{welbl2017sciencqa}, CommonsenseQA~\cite{talmor2019commonsenseqa}, SocialIQA~\cite{sap2019socialiqa}, and OpenbookQA~\cite{mihaylov-etal-2018-suit}.  
In Korean, we use KorQuAD v1~\cite{lim2019korquad1}, KoBEST-Hellaswag, KoBEST-COPA~\cite{kobest}, SNU-Ko-LAMBADA, and SNU-Ko-ARC-Easy~\cite{kim2025thunder}.  

Unlike Dolma~\cite{soldaini2024dolma}, our evaluation suites explicitly include SQuAD (v1) and KorQuAD (v1) for assessing generative question-answering tasks in English and Korean, respectively.  This design ensures that the performance of generative tasks is explicitly considered as a criterion in data curation practices. Full details on the downstream tasks and the evaluation method are described in Appendix~\ref{sec:app-benchmark-detail}.

\subsection{Rethinking Signals from Text Lines}
\label{sec:revisiting-signal-rules}
Lines, delimited by newline characters (\textbackslash n), are often used as the basic unit of filtering in data preprocessing pipelines. Line-level filtering techniques, such as deduplication and punctuation-based rules, are now standard practices in creating datasets derived from CommonCrawl. These methods help identify boilerplate or low-value text, and previous studies have shown that they can improve model performance by reducing redundancy and promoting a more coherent sentence structure. However, relying too strictly on individual signals may lead to the removal of lines that contain important content or structural information, which can compromise the document's integrity and negatively impact the model's performance in downstream tasks. Thus, we revisit two common rules — \textit{line-level deduplication} and \textit{trailing-punctuation filtering} — to assess both their advantages and any unintended consequences. This evaluation supports our recommendation to move beyond these isolated heuristics and instead use document-level signal patterns for more effective filtering.

\paragraph{Line-level deduplication.} 
We revisit line-level deduplication first, which was originally introduced as part of the CCNet pipeline~\cite{wenzek2019ccnet}. This method has been widely adopted in the construction of open-source text datasets \cite{soldaini2024dolma, young2024yi} using CommonCrawl WET. CCNet demonstrated that line-level deduplication effectively removes boilerplate content, such as cookie warnings, extracts unique text from each document, and eliminates repetitive information. This process is believed to enhance the performance of language models~\cite{lee2021deduplicating}.

\begin{figure}[t!]
    \centering
    \includegraphics[width=0.9\columnwidth]{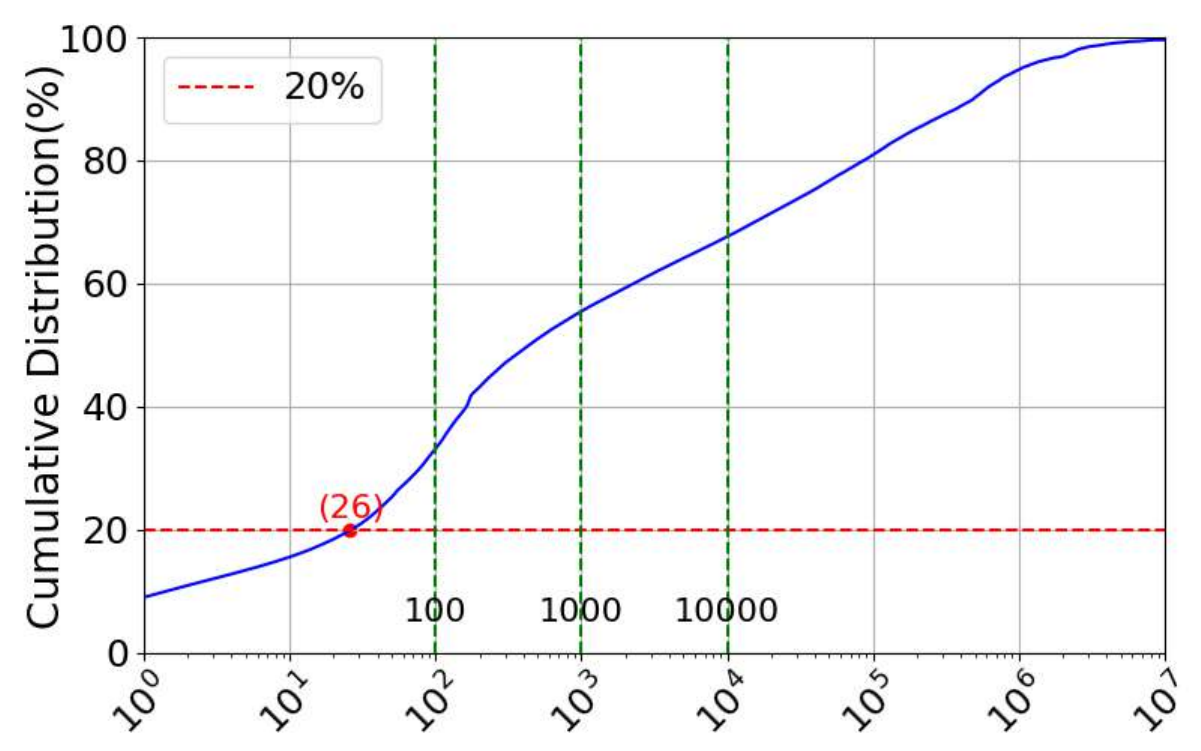}
    \caption{Cumulative distribution of frequency count of text lines (English documents).}
    \label{fig:en_cumulative}
\end{figure}

\begin{figure}[t!]
    \centering
    \includegraphics[width=0.9\columnwidth]{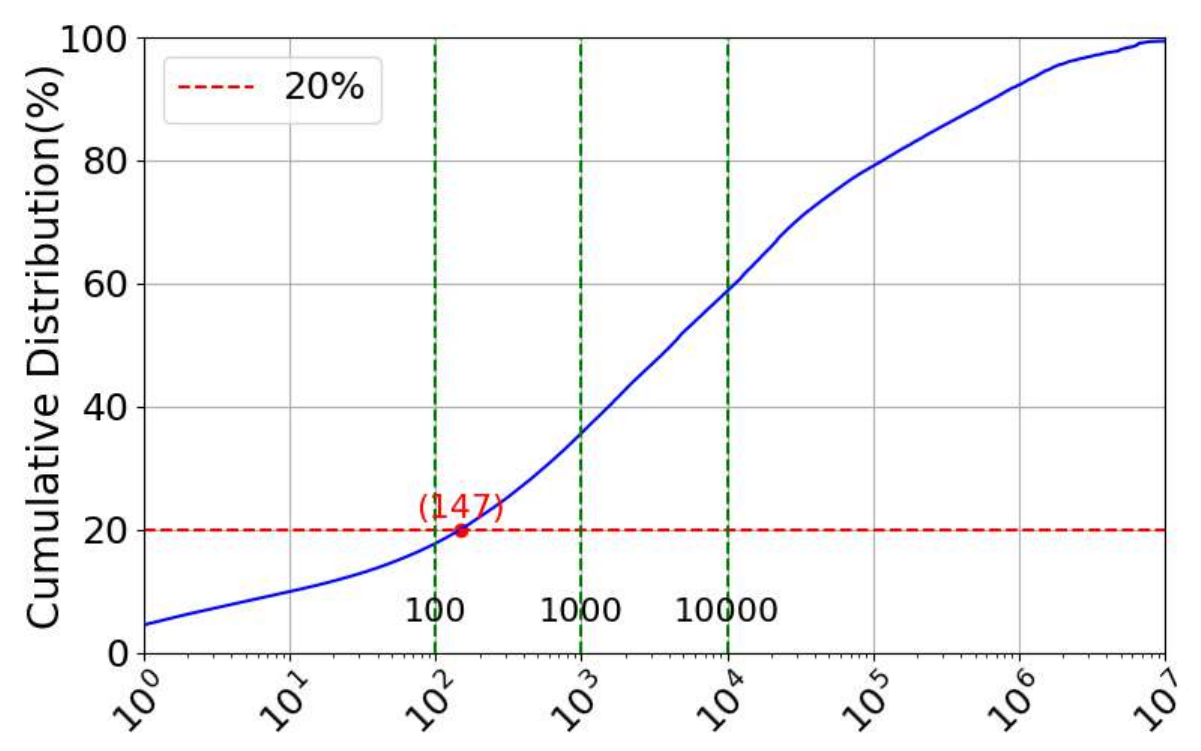}
    \caption{Cumulative distribution of frequency count of text lines (Korean documents).}
    \label{fig:ko_cumulative}
\end{figure}


\newcommand{\tbcell}[1]{#1}

\begin{table*}[t]
\centering
\renewcommand{\arraystretch}{1} 
\resizebox{0.8\textwidth}{!}{
\begin{tabular}{r p{0.65\textwidth} c c}
\toprule
No & \centering Line & Count & Punc. \\
\midrule
1 & \tbcell{There are a lot of reasons ... who has an extensive tee shirt collection.} & 1 & \checkmark \\
2 & \tbcell{However, working with ... challenges. But with ... very doable.} & 1 & \checkmark \\
3 & \tbcell{T-SHIRT QUILTS} & 284 &  \\
4 & \tbcell{Shhhh: Here's the big secret: diligent application ... interfacing.} & 1 & \checkmark \\
5 & \tbcell{Cable tv also helps a lot here.} & 1 & \checkmark \\
6 & \tbcell{This secret, ..., baby clothing, etc. }& 1 & \checkmark \\
7 & \tbcell{You will need:} & 1{,}316 &  \\
8 & \tbcell{A good iron with ... which mists water.} & 1 & \checkmark \\
9 & \tbcell{A thin cotton press cloth (any clean piece of muslin will do).} & 1 & \checkmark \\
\bottomrule
\end{tabular}
}
\vspace{-0.5\baselineskip}
\caption{Examples of line-level signals with frequency counts and punctuation marks. The Punc. column denotes if the line ends with a punctuation mark.}
\label{tab:line_examples}
\end{table*}


Figures~\ref{fig:en_cumulative} and \ref{fig:ko_cumulative} display the cumulative distribution of \textit{count}, which refers to the number of documents in which a specific text line appears. We calculate this across all text lines in the validation set of our baseline corpus (see Appendix~\ref{sec:app-detail-data}).

At the line level, we find that nearly 80\% of English lines are repeated in over 26 documents out of a total of approximately 20 million documents. Korean data shows a similar trend, with more than 80\% of lines appearing in more than 147 documents. However, as noted in OSCAR-2022~\cite{oscar2022}, line-level deduplication can compromise document integrity, particularly when repeated phrases serve structural or semantic roles (such as song lyrics or section headers). Table~\ref{tab:line_examples} illustrates these examples: lines 3 and 7 convey structural cues but would be removed by simple line-level deduplication.

Additionally, we observe that implementing line-level deduplication on English data leads to noticeable performance degradation on SQuAD v1 (see Section~\ref{sec:eval-en}). Overall, while these findings indicate that the number of documents in which a line appears is a useful indicator of its uniqueness, it should not be solely relied upon for filtering purposes.


\paragraph{Trailing-punctuation filtering.} 
Next, we revisit the method of filtering lines based on trailing punctuation marks. This heuristic eliminates lines that do not end with a punctuation mark, such as a period (.), exclamation point (!), question mark (?), and quotation marks (" and '). This approach was first proposed in C4~\cite{raffel2023exploring}, and its effects have been examined through data ablation studies in Dolma~\cite{soldaini2024dolma} and Fineweb~\cite{penedo2024fineweb}. Both of these studies, along with our own experiments on English data, indicate that this filter can enhance performance in downstream tasks. However, consistent with findings from Dolma and Fineweb, we observe that this rule excludes a significant proportion of data—up to 40\% of English tokens and 55\% of Korean tokens.

As illustrated in Table~\ref{tab:line_examples}, text lines containing structural information (e.g., 3 and 7 in Table~\ref{tab:line_examples}) may be discarded when using the trailing-punctuation filter. Furthermore, while our ablation experiments validate improvements for certain multiple-choice tasks, we note a substantial decline in performance for others. For instance, our Korean \base ~model trained with only line-level deduplication achieved an exact match score of 6.5 on the KorQuAD v1 dataset. However, applying the trailing-punctuation filter drastically lowered this score to 0.1. This suggests that although lines ending with punctuation marks are beneficial, many other lines without such marks are also crucial for effective language model training.

\section{\pa Line Filtering \label{sec:palf}}
We now introduce \textit{pattern-aware line filtering} methods that exploit the distribution of text-quality signals across documents instead of treating each line in isolation. While individual signals can offer valuable insights, none should be considered definitive filtering criteria on their own. The raw data example in Table~\ref{tab:line_examples} demonstrates that lines with low-value signals—such as short headers or display elements—often hold structural or semantic significance when viewed in their surrounding context. Based on this observation, we first assign a categorical label to each line derived from text-quality signals. Then, we represent a document as a sequence of these labels and identify characteristic label patterns that differentiate relevant text from irrelevant noise.

\begin{table*}[t]
\centering
\renewcommand{\arraystretch}{1} 
\resizebox{0.8\linewidth}{!}{
\begin{tabular}{c|l|rc|ccc|c}\toprule
No. &Line &Count &Category &(1) &(2) &(3) &Result \\\midrule
1 &Infographic &7556 &R & & & & \\
2 &About &3{,}215{,}635 &R & & & & \\
3 &Contact &3{,}533{,}305 &R & & & & \\
4 &0 &2{,}194{,}233 &R & & & & \\
5 &\tbcell{Best Drop 3 Baseball Bat} &6 &Y & & & & \\
6 &\tbcell{When you are purchasing ... across.If you want ... suits you well.} &1 &G &\checkmark &\checkmark &\checkmark &\checkmark \\ 
\multicolumn{8}{c}{$\vdots$} \\ 
12 &\tbcell{Drop three ... to wood bats. This is ... the bat.} &1 &G &\checkmark &\checkmark &\checkmark &\checkmark \\
13 &\tbcell{Numerical Difference} &1 &G &\checkmark &\checkmark &\checkmark &\checkmark \\
14 &\tbcell{For a baseball bat to be regarded ... stated of three.} &1 &G &\checkmark &\checkmark &\checkmark &\checkmark \\
15 &\tbcell{Handle} &1{,}608 &R & & &\checkmark &\checkmark \\
16 &\tbcell{The handle of ... field. Choose ... strength. The ... time.} &1 &G &\checkmark &\checkmark &\checkmark &\checkmark \\
17 &\tbcell{Other Considerations} &391 &Y & & \checkmark&\checkmark &\checkmark \\
18 &\tbcell{Determine the diameter of the barrel} &1 &G &\checkmark &\checkmark &\checkmark &\checkmark \\
19 &\tbcell{Look at the weight of the bat} &1 &G &\checkmark &\checkmark &\checkmark &\checkmark \\
\bottomrule
\end{tabular}
}
\vspace{-0.5\baselineskip}
\caption{ An example of applying the pattern-aware line-level deduplication (PLD). }
\label{tab:pld-example}
\end{table*}

\paragraph{Pattern-aware line-level deduplication (\pld).}
We enhance line-level deduplication by exploiting frequency count signals to determine whether a line is unique to a document or highly repetitive across multiple documents. Each line is classified into one of three categories based on \textit{the number of documents} in which it appears. First, we process all documents in a document set, which consists of approximately 20 million documents. Each CommonCrawl snapshot typically contains between 15 million and 25 million Korean documents, collected over a specific time period, usually spanning one to two months. Next, we construct a hashmap to track the number of documents each hashed line appeared. For the implementation details, see Appendix \ref{sec:app-detail-dedup}.

For each line in a document, we assign a category: Red (highly repetitive), Yellow (undecidable), or Green (distinctive) based on its frequency. A line is categorized as highly repetitive (Red) if its count exceeds 1,000 for English or 50 for Korean. A line is classified as undecidable (Yellow) if its count is greater than 1 for English or more than 3 for Korean. All other lines are classified as distinctive (Green). We determine these frequency thresholds by testing various configurations and performing data ablations using \sm ~models. As shown in Figures~\ref{fig:en_cumulative} and \ref{fig:ko_cumulative}, the count distribution differs by language, so we establish the thresholds for each language. For details on the threshold tuning procedure, please refer to Appendix \ref{sec:tuning-pald}.

With the results of the categorization, we focus on retaining sequences of lines that meet one of the following criteria: 
\begin{itemize}
\setlength{\itemsep}{0pt}
\setlength{\parskip}{0pt}     
\setlength{\parsep}{0pt}
\item Two or more consecutive distinctive (Green) lines, 
\item One or more undecidable (Yellow) lines embedded between runs of distinctive lines
\item Up to three undecidable (Yellow) or highly repetitive (Red) lines embedded between runs of distinctive (Green) lines. 
\end{itemize}
We have observed that a distinctive line appearing in isolation is typically unrelated or erroneous, making it irrelevant for downstream tasks. Therefore, unlike conventional line-level deduplication — where we retain a single line only if its count is one — we remove isolated distinctive lines unless they match one of the predefined patterns above. In addition, the retained subsequence must start and end with at least two consecutive distinctive lines, ensuring that isolated noise is excluded.

Table~\ref{tab:pld-example} shows an example of applying PLD. A checkmark in columns (1), (2), and (3) indicates that the line matches the corresponding pattern used in PLD. A checkmark in the Result column signifies that the line will be preserved. We remove irrelevant text (boilerplate) found in lines 1–5 while preserving lines 15 and 17, which would otherwise be eliminated under traditional line-level deduplication.

For implementation, we encode the sequence of classifications as a string over the alphabet \{r, y, g\}, which correspond to Red, Yellow, and Green. We then utilize regular expressions to identify these patterns and retain only the text lines that fall within the matched spans. Specifically, the regex patterns used are as follows: 
\begin{itemize}
\setlength{\itemsep}{0pt}
\setlength{\parskip}{0pt}     
\setlength{\parsep}{0pt}
\item \verb|g{2,}|: this captures consecutive distinctive lines. 
\item \verb|g{2,}(y+g{1,})+g|: this captures undecidable stretches that are bounded by distinctive lines. 
\item \verb|g{2,}([yr]{0,3}g{1,})+g|: this captures short stretches (of at most three lines) of undecidable or highly repetitive lines that are also bounded by distinctive lines.
\end{itemize}

\paragraph{Pattern-aware trailing-punctuation filtering (\ptf).}
We also enhance the trailing-punctuation filtering by using punctuation signals to determine whether a line is likely part of a complete sentence. For instance, Lines 3 and 7 in Table~\ref{tab:line_examples} illustrate that text lines without trailing punctuation marks can still serve as important structural cues within a document. We notice that such lines typically occur between lines that do end with punctuation marks. Based on this observation, we categorize each line into two groups according to the presence of a trailing punctuation mark: with (Green) and without (Red) trailing punctuation marks. Using this categorization, we retain sequences that meet both of the following criteria: 
\begin{itemize}
\setlength{\itemsep}{0pt}
\setlength{\parskip}{0pt}     
\setlength{\parsep}{0pt}
\item One or more consecutive lines ending with punctuation marks
\item Up to $k$ consecutive lines without trailing punctuation marks that are positioned between lines that end with punctuation marks.  
\end{itemize}

\begin{table*}[t]
\centering
\renewcommand{\arraystretch}{1} 
\resizebox{0.88\linewidth}{!}{
\begin{tabular}{c|l|cc|cc|c}\toprule
No. &Line &Punc. &Category &(1) &(2) &Result \\\midrule
1 &\tbcell{For a baseball bat to be regarded ... stated of three.} &\checkmark &G &\checkmark &\checkmark &\checkmark \\
2 &Handle & &R & &\checkmark &\checkmark \\
3 &\tbcell{The handle of ... field. Choose ... strength. The ... time.} &\checkmark &G &\checkmark &\checkmark &\checkmark \\
4 &Other Considerations & &R & &\checkmark &\checkmark \\
5 &Determine the diameter of the barrel & &R & &\checkmark &\checkmark \\
6 &Look at the weight of the bat & &R & &\checkmark &\checkmark \\
7 &Does the bat come with a warranty? &\checkmark &G &\checkmark &\checkmark &\checkmark \\
\bottomrule
\end{tabular}
}
\vspace{-0.5\baselineskip}
\caption{An example of applying pattern-aware trailing-punctuation filtering (PTF). 
}
\label{tab:ptf-example}
\end{table*}

Table~\ref{tab:ptf-example} shows an example of applying PTF. A checkmark in columns (1) and (2) indicates that the line corresponds to the specific pattern used in PTF. A checkmark in the Result column signifies that the line will be retained. Structural cues in lines 2 and 4, along with relevant text in lines 5 and 6, are preserved with PTF; otherwise, they would be discarded by the traditional trailing-punctuation filtering method.

For implementation, we encode the sequence of classifications as a string over the alphabet \{r, g\}, corresponding to Red (without punctuation marks) and Green (with punctuation marks). We then use regular expressions to match these patterns and retain only the text lines that belong to matched spans. 
Concretely, the regex patterns are:  
\{ \verb|"g+"|, \verb|"g+r{,k}g+"|\}.  
The first captures runs of sentence-like lines ending with punctuation; the second allows short stretches (at most $k$ lines) without punctuation when they are embedded between sentence-like lines. 

We determine the value of $k$ by testing multiple variants and performing data ablations with \sm ~models. For English, we use $k=3$ while for Korean, we use $k=15$.
See Appendix \ref{sec:tuning-patp} for details of the threshold tuning procedure.

\section{Evaluation}
\label{sec:evaluation}

In accordance with the data ablation settings outlined in Section~\ref{sec:methodology}, we train the \base ~models using the processed datasets. To reduce variability in the filtering decisions, each configuration is trained twice. In our results, \textit{LD} refers to line-level deduplication, and \textit{TF} stands for the trailing-punctuation filter. Additionally, \textit{\pld} and \textit{\ptf} represent their pattern-aware counterparts, our proposed methods. For implementation details regarding LD, please refer to Appendix~\ref{sec:app-detail-dedup}. When applying the trailing-punctuation filter (TF), we eliminate all lines that do not end with a punctuation mark (., ?, !, ", or ').

\begin{table*}[t!]
\centering
\begin{minipage}{\linewidth}
\centering
\resizebox{0.9\linewidth}{!}{%
\begin{tabular}{@{}c|cccccccc|cc@{}}
\toprule
Method & ARC(E) & CSQA & Hellaswag & OBQA & PIQA & SciQ & SIQA & SQuAD (v1) & Mean (w.o. SQuAD) & Mean \\
\midrule\midrule
Baseline     & 46.7 & 38.2 & 36.9 & 31.5 & 66.4 & 64.0 & 42.6 & 7.3 & 46.6 & 41.7 \\
LD           & 52.1 & 44.5 & 44.9 & \textbf{35.0} & 69.6 & \textbf{73.2} & \textbf{44.1} & 0.1 & 51.3 & 44.9 \\
LD + TF      & 43.4 & 30.6 & \textbf{47.8} & \ul{33.6} & \ul{71.0} & 69.0 & 43.1 & 0.0 & 48.4 & 42.3 \\
\midrule
\pld         & \ul{54.0} & \ul{47.3} & 45.7 & 33.0 & \ul{71.0} & 72.2 & \ul{43.5} & \textbf{14.9} & \ul{52.4} & \ul{47.7} \\
\pld+\ptf    & \textbf{54.8} & \textbf{48.8} & \ul{47.7} & 33.2 & \textbf{71.4} & \ul{73.1} & 43.4 & \ul{12.5} & \textbf{53.2} & \textbf{48.1} \\
\bottomrule
\end{tabular}}\
\vspace{-0.5\baselineskip}
\caption{Experimental results on English data (\textbf{base(1B)}). The best scores are in bold, and the second-best scores are underlined.}
\label{tab:exp-en-base}
\end{minipage}
\\
\vspace{1em}
\begin{minipage}{\linewidth}
\centering
\resizebox{0.85\linewidth}{!}{%
\begin{tabular}{@{}c|ccccc|cc@{}}
\toprule
Method & 
\begin{tabular}{c}
KoBEST-\\
COPA 
\end{tabular} & 
\begin{tabular}{c}
KoBEST-\\
Hellaswag 
\end{tabular} & 
\begin{tabular}{c}
SNU-Ko-\\
ARC(E) 
\end{tabular} & 
\begin{tabular}{c}
SNU-Ko-\\
Lambada 
\end{tabular} & 
\begin{tabular}{c}
KorQuAD \\
v1 \end{tabular} 
& 
\begin{tabular}{c}
Mean without\\
KorQuAD 
\end{tabular} & 
Mean \\
\midrule\midrule
Baseline & 60.6 & 42.1 & 37.7 & 74.4 & 0.2 & 53.7 & 43.0 \\
LD       & 68.8 & 50.3 & 45.9 & 81.0 & \ul{6.5} & 61.5 & \ul{50.5} \\
LD + TF  & \ul{70.7} & \ul{50.7} & 45.7 & \textbf{82.1} & 0.1 & \ul{62.3} & 49.9 \\
\midrule
\pld     & 69.6 & \ul{50.7} & \ul{46.2} & \ul{81.9} & \textbf{7.9} & 62.1 & \textbf{51.3} \\
\pld+\ptf & \textbf{70.8} & \textbf{53.0} & \textbf{48.4} & 80.6 & 3.9 & \textbf{63.2} & \textbf{51.3} \\
\bottomrule
\end{tabular}}
\vspace{-0.5\baselineskip}
\caption{Experimental results on Korean data (\textbf{base(1B)}). The best scores are in bold, and the second-best scores are underlined.}
\label{tab:exp-ko-base}
\end{minipage}

\end{table*}

\subsection{English Dataset}
\label{sec:eval-en}

Table~\ref{tab:exp-en-base} presents the performance of \base ~models on various downstream tasks using different preprocessed versions of the English dataset. We observe that employing pattern-aware line-filtering techniques enhances overall downstream performance. Our filtering approach enables \base  ~models to effectively tackle the SQuAD v1 task, while also achieving accuracy improvements on multiple-choice tasks. Specifically, our method shows accuracy gains on 3 out of 7 tasks, while still maintaining competitive performance on the other tasks. As a result, our method yields a mean accuracy across multiple-choice benchmarks that exceeds those of the pattern-agnostic data filtering methods.

\subsection{Korean Dataset}
\label{sec:eval-ko}

Table~\ref{tab:exp-ko-base} presents the performance on downstream tasks for \base ~models trained using various preprocessed versions of the Korean dataset. Our findings indicate that pattern-aware filtering consistently surpasses its pattern-agnostic counterparts. Notably, substituting LD with \pld results in improved accuracy across all tasks. Additionally, when compared to the Dolma-style pipeline~\cite{soldaini2024dolma} that applies TF following LD, replacing both with their pattern-aware versions leads to accuracy improvements in 4 out of 5 tasks. When comparing overall accuracy across all benchmarks, the application of either PLD alone or both PLD and PTF produces the highest average scores. Specifically, when evaluating the average accuracy across all benchmarks but KorQuAD, the combination of both PLD and PTF achieves the best result, offering a significant advantage over the Dolma-style pipeline.

\begin{table}[t!]
\centering
\setlength{\tabcolsep}{3pt}
\renewcommand{\arraystretch}{0.9}
\resizebox{\columnwidth}{!}{%
\begin{tabular}{@{}l|rrr|rr@{}}
\toprule
& Baseline & LD & LD+TF & \pld & \pld+\ptf \\ \midrule \midrule
English & 73,216 & 20,592 & 12,883 & 18,702 & 13,238 \\
Korean  & 1,010  &   206  &    93  &    158 &    102  \\ 
\bottomrule
\end{tabular}%
}
\vspace{-0.5\baselineskip}
\caption{Token counts in billions (B) for English and Korean datasets under different filtering configurations. 
\label{tab:filtered-data-detail}}
\end{table}

\subsection{Token Counts}
\label{sec:eval-ko}
We track the number of tokens remaining after each filtering step. For the English dataset, we randomly sample 1\% of the training corpus, tokenize it, and then estimate the total token count by multiplying the sample size by 100. The token count is measured using the LLaMA3 tokenizer~\cite{touvron2023llama}. For the Korean dataset, we tokenize the entire training corpus and count the tokens directly. The token count for this dataset is measured with the Exaone 4.0 tokenizer~\cite{exaone-4.0}. We present the number of tokens for all the preprocessed training datasets in Table~\ref{tab:filtered-data-detail}. 

Interestingly, the pattern-aware line-level deduplication technique removes more tokens than the pattern-agnostic method. This is due to our design choice to eliminate isolated, distinctive lines. Conversely, the pattern-aware trailing-punctuation filtering removes fewer tokens compared to its pattern-agnostic equivalent. This outcome is expected because the set of lines retained by the pattern-aware filter includes all those retained by the pattern-agnostic filter.

\subsection{Cross-Language Comparison}
As discussed in Section~\ref{sec:revisiting-signal-rules}, the factors contributing to improvements or declines in accuracy vary between languages and filtering methods. For instance, in English, applying line-level deduplication results in decreased accuracy on the SQuAD v1 dataset, while in Korean, it enhances performance. This indicates that filtering decisions should be tailored to each language. Based on this insight, we adjusted the thresholds for our pattern-aware filtering rules separately for English and Korean.

We consider the benchmark pairs Hellaswag vs. KoBEST-Hellaswag, ARC-Easy vs. SNU-Ko-ARC-Easy, and SQuAD v1 vs. KorQuAD v1 to be analogous due to their similar structures and domains. Our findings indicate that when pattern-aware filtering enhances accuracy in one language, it often has a positive effect on the corresponding benchmark in the other language as well. For instance, applying PTF after PLD increases the accuracy of Hellaswag by 2.0 points, while its Korean counterpart improves by 2.33 points.

On the other hand, we observe that applying \ptf harms the performance of short-form generative question answering tasks, particularly in SQuAD v1 and KorQuAD v1. While \pld enhances accuracy compared to its pattern-agnostic version, the subsequent application of \ptf results in a reduction in accuracy for both languages. 
These findings indicate a trade-off caused by PTF: although it improves performance on multiple-choice benchmarks, it adversely affects generative question-answering tasks.
\label{sec:impact}

\section{Conclusions}
\label{sec:conclusion}
In this study, we revisit common line-level filtering heuristics used in creating large-scale web corpora and highlight their limitations, particularly concerning generative question-answering (QA) tasks. We introduce pattern-aware line filtering methods, which takes into account the distributional patterns of filtering signals across documents rather than treating each line in isolation. Through ablation studies conducted on both English and Korean datasets, we demonstrate that pattern-aware rules consistently outperform their pattern-agnostic counterparts. 
Our findings indicate that filtering decisions should consider not only individual line signals but also the broader context of the document. We believe this proposed methodology can be applied in various settings where document integrity and detailed signal interactions are crucial for model performance, such as in domain-specific corpus construction and in continual pretraining pipelines. 

\section*{Limitations 
\label{sec:limitations}}

We do not investigate the impact of other steps in the data construction pipeline, such as model-based filtering or document-level deduplication.  

Although we suggest promising parameters for the filtering process, it is uncertain whether parameters tuned for one language are optimal for all languages. Future work should examine this more thoroughly.  

While we invested substantial computational resources in our experiments, we were only able to perform a limited hyperparameter search and repeated trials are limited to twice per experiment. It remains unclear whether marginal differences caused by random fluctuations may have led to suboptimal hyperparameter choices.  

Finally, we have assessed the effectiveness of our technique only in English and Korean. Its applicability to other languages remains to be explored.  

\section*{Ethics Statement \label{sec:ethics}}

We utilize publicly available web data as the source data and we properly utilized the data following the restrictions stated in the license. Hence, our data collection does no have legal concerns.

We will not distribute the processed data. Therefore, there are no copyright issues nor safety concerns.



\section*{Acknowledgements \label{sec:ack}}

This work was partially supported by the National Research Foundation of Korea (NRF) under Grant No. RS-2023-00222663 (Center for Optimizing Hyperscale AI Models and Platforms), and by the Institute for Information and Communications Technology Promotion (IITP) under Grant No. 2018-0-00581 (CUDA Programming Environment for FPGA Clusters) and No. RS-2025-02304554 (Efficient and Scalable Framework for AI Heterogeneous Cluster Systems), all funded by the Ministry of Science and ICT (MSIT) of Korea. It was also partially supported by the Korea Health Industry Development Institute (KHIDI) under Grant No. RS-2025-25454559 (Frailty Risk Assessment and Intervention Leveraging Multimodal Intelligence for Networked Deployment in Community Care), funded by the Ministry of Health and Welfare (MOHW) of Korea. Additional support was provided by the BK21 Plus Program for Innovative Data Science Talent Education (Department of Data Science, Seoul National University, No. 5199990914569) and the BK21 FOUR Program for Intelligent Computing (Department of Computer Science and Engineering, Seoul National University, No. 4199990214639), both funded by the Ministry of Education (MOE) of Korea. This work was also partially supported by the Artificial Intelligence Industrial Convergence Cluster Development Project, funded by the MSIT and Gwangju Metropolitan City. Research facilities were provided by the Institute of Computer Technology (ICT) at Seoul National University.


\bibliography{anthology,custom}

\clearpage
\appendix
\section{Experiment Details \label{sec:app-experiment-detail}}

\subsection{Training Data\label{sec:app-detail-data}}

\paragraph{Snapshots} A CommonCrawl snapshot refers to a collection of documents gathered within a specific period, typically spanning one to two months. We select the first five snapshots from each year between 2019 and 2022. Within a snapshot, documents are evenly distributed across 56,000 to 80,000 shards. We treat shards as the minimal unit of data control. For each language, we perform shard sampling only once at the beginning of the experiment setup, and consistently use the same sampled shards for all subsequent experiments in that language. This design ensures that experimental conditions are comparable across experiments and that all randomly selected documents originate from the same raw data source.

\paragraph{Korean Data} We extract Korean documents in which at least 10\% of the characters are Hangul (가–힣). We allocate 1\% of the total shards as the validation split, and use the remainder as training data.

\paragraph{English Data} We extract English documents that achieve an English score of 0.65 or higher, as classified by the FastText\cite{joulin2016fasttext} language identifier. From this set, we sample 0.2\% of the shards for validation, and 1\% of the shards for training.

\subsection{Model and Training\label{sec:app-detail-model}}

We follow LLaMA 3.2 1B architecture while modifying the number of layers, hidden dimensions, and number of attention heads to adjust the model size. Table summarizes the details regarding the size of the model.

Both model scales were trained with a sequence length of 2{,}048 and 288 examples per iteration, resulting in a total of 589{,}824 tokens per iteration. Both scales used the AdamW\cite{adamw} optimizer with $\beta_1 = 0.9$ and $\beta_2 = 0.95$. We set the target learning rate to $4 \cdot 10^{-4}$ for the small (900M) models and $2 \cdot 10^{-4}$ for the base (1B) models. The learning rate was linearly warmed up over 1,000 steps, starting from 10\% of the target value, and then kept constant once the target learning rate was reached. The small (900M) models were trained for 34,000 iterations, corresponding to 20B tokens, while the base (1B) models were trained for 52,000 iterations, corresponding to 30B tokens. We employed DeepSpeed\cite{rasley2020deepspeed} ZeRO-1 to distribute optimizer states across multiple GPUs.

\subsection{Computation Costs for Training Models\label{sec:app-computation-train}}

Our training infrastructure consists of 11 GPU Nodes, each node providing 6 RTX 4090 GPUs. All GPU Nodes are interconnected with NVIDIA Mellanox Infiniband HDR Network, providing 200 Gbps interconnection per node. Table\ref{tab:gpu-costs} shows total computation costs for training models.
\begin{table}[htbp]
\centering
\setlength{\tabcolsep}{3pt}
\renewcommand{\arraystretch}{0.9}
\resizebox{\columnwidth}{!}{%
\begin{tabular}{@{}l|cc|c@{}}
\toprule
Content &Small(900M) &Base(1B) &Total \\ \midrule \midrule
Number of Nodes &1 &4 &- \\
Number of GPUs &6 &24 &- \\
Training Time(Hours) &48 &48 &- \\
GPU Hours per Trial &288 &1,152 &- \\
Number of Training Trials &104 &20 &124 \\ \midrule \midrule
Total GPU Hours &29,952 &23,040 &52,992 \\
\bottomrule
\end{tabular}}
\caption{Computation Costs for Training Models}\label{tab:gpu-costs}
\end{table}

\subsection{Implementation of LD and PLD \label{sec:app-detail-dedup}}

We first define the document set for computing the count. We set the document set size to be on the order of 20 million documents, because each snapshot contains between 15 million and 25 million Korean documents.

\paragraph{Definition of Document Set in English} 
While CCNet\cite{wenzek2019ccnet} performs line-level deduplication first in order to improve language identification accuracy, 
we instead begin with language identification to prevent count information from being affected by documents in other languages. 
To construct document sets of approximately 20 million English documents, we group shards accordingly. 
Since each shard contains on average 20{,}000 documents, we define a document set as 
$S_i = S[i \cdot 1000:(i+1) \cdot 1000]$, 
where $S$ denotes the ordered list of shards in a snapshot and $S_i$ the $i$-th document set. 
Because documents crawled from the same domain on the same date are placed within the same shard or adjacent shards, 
this grouping facilitates the detection of boilerplate content from the same website using count signals.

\paragraph{Definition of Document Set in Korean} 
For Korean, we first perform language identification by detecting Hangul characters. 
This design choice is motivated by the fact that Korean is the only high/mid-resource language that exclusively uses Hangul. 
After identifying Korean documents, we define one document set per snapshot.

\paragraph{Counting the Occurrence in a Document Set} 
For hash computation, we first normalize each line, and then compute its SHA1 hash for comparison. 
In the normalization step, we replace all upper case alphabets with lowercase alphabets, digits with 0.
We also normalize Unicode punctuation mark characters to ASCII counterparts and blank characters to ASCII space characters.
We maintain a hashmap where each hash is associated with the number of occurrences, stored in \texttt{uint16} format. 
Our implementation accounts for arithmetic overflow: if the count exceeds the maximum value, 
the stored value is saturated at the maximum. 
In an earlier implementation, we used \texttt{uint32} format, which allowed us to analyze the full distribution of counts, 
as illustrated in Figure~\ref{fig:en_cumulative} and Figure~\ref{fig:ko_cumulative}. 
The count information for each line is obtained by looking up its corresponding hash in the hashmap. 
For line-level deduplication, we discard all lines with a count larger than 1.
We utilize this count signal for pattern-aware line-level deduplication.

\section{Threshold Tuning \label{sec:tuning}}

\subsection{Pattern-Aware Line-Level Deduplication(PLD) \label{sec:tuning-pald}}

As described in Section \ref{sec:palf}, we classify lines by their occurrence counts and use these categorizations to identify patterns corresponding to meaningful text. Pattern-aware line-level deduplication categorizes each line into one of three groups: \textit{highly repetitive} if its count exceeds $r$, \textit{undecidable} if its count exceeds $g$, and \textit{distinctive} otherwise. In this section, we describe how we determined the categorization thresholds of $r=1000, g=1$ for English and $r=50, g=3$ for Korean. We conduct data ablation experiments with \textbf{small (900M)} models to make these decisions. For each configuration, we train the model twice, compute the mean accuracy across the two runs, and select the thresholds that yield the highest mean accuracy across all benchmarks.

\paragraph{English}

\begin{table}[htbp]
\centering
\setlength{\tabcolsep}{4pt}
\renewcommand{\arraystretch}{0.9}
\resizebox{\columnwidth}{!}{%
\begin{tabular}{@{}c|cccccc@{}}
\toprule
\multirow{2}{*}{$r$} & \multicolumn{6}{c}{$g$} \\ 
\cmidrule(l){2-7}
 & 1 & 3 & 5 & 10 & 15 & 20 \\
\midrule\midrule
50   & 46.0 & 45.8 & 45.6 & 44.6 & 45.3 & 44.5 \\
100  & 45.5 & 45.6 & 45.1 & 45.1 & 45.1 & 44.9 \\
500  & 45.3 & 45.6 & 45.6 &   -   &   -   &   -   \\
1000 & \textbf{46.1} & 45.7 & \ul{46.0} & 45.1 & 44.7 & 44.3 \\
2500 & 45.8 & 45.6 & 45.3 &   -   &   -   &   -   \\
5000 & 46.0 & 44.7 & 44.5 &   -   &   -   &   -   \\
\bottomrule
\end{tabular}}
\caption{Mean accuracy for different values of $r$ and $g$ (English). The best scores are in bold, and the second-best scores are underlined.}
\label{tab:tuning-en-pald}
\end{table}

\begin{table*}[t]
\centering
\resizebox{\textwidth}{!}{%
\begin{tabular}{@{}c|cccccccc|cc@{}}
\toprule
Method & ARC(E) & CSQA & Hellaswag & OBQA & PIQA & SciQ & SIQA & SQuAD (v1) & Mean (w.o. SQuAD) & Mean \\
\midrule\midrule
$r=1000, g=1$   & \textbf{51.5} & 44.2 & \textbf{42.4} & \ul{32.6} & 68.8 & 70.3 & 42.9 & 16.0 & \ul{50.4} & \textbf{46.1} \\
$r=1000, g=5$   & 50.1 & 43.4 & 41.5 & 31.6 & \textbf{69.2} & 70.4 & \textbf{44.2} & \textbf{18.0} & 50.0 & \ul{46.0} \\
$r=50, g=1$     & 51.2 & \textbf{45.7} & \ul{42.0} & 32.0 & 68.7 & \ul{71.0} & \ul{43.4} & 14.1 & \textbf{50.6} & \ul{46.0} \\
$r=5000, g=1$   & \ul{51.4} & 43.1 & \ul{42.0} & 32.0 & 68.8 & 70.7 & 43.2 & \ul{17.0} & 50.2 & 46.0 \\
$r=5000, g=3$   & 49.7 & \ul{44.4} & 41.9 & \textbf{33.0} & \ul{68.9} & \textbf{71.2} & 43.3 & 14.1 & 50.4 & 45.8 \\
\bottomrule
\end{tabular}}
\caption{Per-benchmark accuracy for Top-5 Pattern-Aware Line-Level Deduplication configurations on English data. The best scores are in bold, and the second-best scores are underlined.}
\label{tab:tuning-en-pald-detail}
\end{table*}

We first perform a grid search with $r \in \{50, 100, 1000\}$ and $g \in \{1, 3, 5, 10, 15, 20\}$. Since values of $g \in \{10, 15, 20\}$ are unpromising, we change the search space to $r \in \{50, 100, 500, 1000, 2500, 5000\}$ and $g \in \{1, 3, 5\}$. Table \ref{tab:tuning-en-pald} reports the mean accuracy of models trained with each configuration. Based on these results, we select $r=1000, g=1$. Table \ref{tab:tuning-en-pald-detail} shows the downstream task performance of the top five configurations. While the differences in mean accuracy are small, we note that the third configuration performs particularly well on multiple-choice tasks but shows lower accuracy on SQuAD. This suggests that threshold selection can be tailored to optimize specific benchmark types when desired.

\paragraph{Korean}

\begin{table}[htbp]
\centering
\setlength{\tabcolsep}{4pt}
\renewcommand{\arraystretch}{0.9}
\resizebox{0.5\columnwidth}{!}{%
\begin{tabular}{@{}c|ccc@{}}
\toprule
\multirow{2}{*}{$r$} & \multicolumn{3}{c}{$g$} \\
\cmidrule(l){2-4}
 & 1 & 3 & 5 \\
\midrule\midrule
50   & 51.5 & \textbf{52.4} & 50.2 \\
100  & 51.3 & 51.1 & \ul{52.3} \\
500  & 50.1 & 52.1 & 51.2 \\
1000 & 51.1 & 50.2 & 50.7 \\
2500 & 51.5 & 51.1 & 51.2 \\
5000 & 50.8 & 50.3 & 50.2 \\
\bottomrule
\end{tabular}}
\caption{Mean accuracy for different values of $r$ and $g$ (Korean). The best scores are in bold, and the second-best scores are underlined.}
\label{tab:tuning-ko-pald}
\end{table}

\begin{table*}[htb]
\centering
\resizebox{\textwidth}{!}{%
\begin{tabular}{@{}c|ccccc|cc@{}}
\toprule
Method & KoBEST-COPA & KoBEST-Hellaswag & KorQuAD (v1) & SNU-Ko ARC(E) & SNU-Ko Lambada & Mean (w.o. KorQuAD) & Mean \\
\midrule\midrule
$r=50, g=3$     & 68.1 & \textbf{51.0} & \textbf{18.2} & 44.3 & 80.7 & \textbf{61.0} & \textbf{52.4} \\
$r=100, g=5$    & 67.8 & 50.1 & \ul{17.6} & 44.5 & \textbf{81.6} & \textbf{61.0} & \ul{52.3} \\
$r=500, g=3$    & 67.6 & 50.0 & 17.4 & \ul{44.6} & 80.7 & \ul{60.7} & 52.1 \\
$r=2500, g=1$   & \ul{68.4} & 47.9 & 16.6 & \textbf{44.7} & 80.0 & 60.2 & 51.5 \\
$r=50, g=1$     & \textbf{69.4} & \ul{50.8} & 10.7 & 45.2 & \ul{81.4} & 61.7 & 51.5 \\
\bottomrule
\end{tabular}}
\caption{Per-benchmark accuracy for Top-5 Pattern-Aware Line-Level Deduplication configurations on Korean data. Best scores are in bold, and second-best scores are underlined.}
\label{tab:tuning-ko-pald-detail}
\end{table*}

Following the findings from English experiments, we search over $r \in \{50, 100, 500, 1000, 2500, 5000\}$ and $g \in \{1, 3, 5\}$. We find that $r=50$ and $g=3$ yield the highest mean accuracy. Table \ref{tab:tuning-ko-pald} reports the aggregated results, and Table \ref{tab:tuning-ko-pald-detail} presents the downstream task performance of the top five configurations. We observe that the fifth configuration in Table~\ref{tab:tuning-ko-pald-detail} excels in multiple-choice tasks but performs significantly worse on KorQuAD. This indicates that threshold selection may also be adapted to prioritize particular task types when needed.

\subsection{Pattern-aware Trailing-Punctuation Filtering(PTF) \label{sec:tuning-patp}}

As described in Section \ref{sec:palf}, this filter checks whether lines end with a punctuation mark and allows a sequence of up to $k$ non-punctuated lines if they are enclosed by lines ending with punctuation. We set $k=3$ for English and $k=15$ for Korean. In this section, we describe how these thresholds were chosen. We again perform data ablation with \textbf{small (900M)} models. For each configuration, we train the model twice and select the threshold that achieves the highest mean accuracy across benchmarks.

\paragraph{English}

\begin{table*}[htb]
\centering
\resizebox{\textwidth}{!}{%
\begin{tabular}{@{}c|cccccccc|cc@{}}
\toprule
Method & ARC(E) & CSQA & Hellaswag & OBQA & PIQA & SciQ & SIQA & SQuAD (v1) & Mean (w.o. SQuAD) & Mean \\
\midrule\midrule
$k=1$   & 50.1 & 43.1 & \ul{43.9} & \ul{33.3} & \textbf{70.3} & 69.0 & 43.3 & 0.8 & 50.4 & 44.2 \\
$k=3$   & \textbf{52.1} & \textbf{46.2} & \textbf{44.1} & \textbf{34.0} & 69.7 & \textbf{71.1} & 42.9 & 10.8 & \textbf{51.4} & \textbf{46.4} \\
$k=5$   & \ul{52.0} & 44.6 & 43.6 & 31.9 & \ul{69.9} & 68.9 & 43.1 & 7.7 & 50.6 & 45.2 \\
$k=7$   & 51.4 & 44.4 & 43.4 & 32.3 & 69.7 & \ul{70.6} & \ul{43.5} & 12.5 & 50.7 & 46.0 \\
$k=10$  & 51.8 & \ul{45.5} & 43.6 & 33.2 & 69.4 & \ul{70.6} & 43.3 & \ul{12.6} & \ul{51.1} & \ul{46.2} \\
$k=15$  & \textbf{52.1} & 44.1 & 43.3 & 31.9 & 69.2 & 70.4 & \textbf{43.6} & \textbf{14.2} & 50.7 & 46.1 \\
\bottomrule
\end{tabular}}
\caption{Per-benchmark accuracy for different hyperparameter settings ($k$) on English data. The best scores are in bold, and the second-best scores are underlined.}
\label{tab:tuning-en-patp-detail}
\end{table*}

After applying pattern-aware line-level deduplication with $r=1000$ and $g=1$, we search over $k \in \{1, 3, 5, 7, 10, 15\}$. We find that $k=3$ yields the best average accuracy. Table \ref{tab:tuning-en-patp-detail} summarizes the results. We note that the accuracy on SQuAD increases as $k$ increases. However, even with larger $k$, the SQuAD performance does not reach the accuracy observed without this filtering process. This indicates that applying this filter improves overall accuracy on multiple-choice tasks but may sacrifice performance on generative QA benchmarks.

\paragraph{Korean}

\begin{table*}[htb]
\centering
\resizebox{\textwidth}{!}{%
\begin{tabular}{@{}c|ccccc|cc@{}}
\toprule
Method & KoBEST-COPA & KoBEST-Hellaswag & KorQuAD (v1) & SNU-Ko ARC(E) & SNU-Ko Lambada & Mean (w.o. KorQuAD) & Mean \\
\midrule\midrule
$k=1$   & 68.9 & 50.5 & 10.7 & 44.9 & 81.3 & 61.4 & 51.3 \\
$k=3$   & \textbf{70.2} & \ul{50.8} & \ul{14.7} & 45.2 & 81.0 & 61.8 & \ul{52.4} \\
$k=5$   & 69.6 & \textbf{52.1} & 13.9 & 45.7 & 81.4 & \textbf{62.2} & 52.5 \\
$k=7$   & \ul{69.9} & 50.0 & 10.3 & 44.9 & \textbf{82.0} & 61.7 & 51.4 \\
$k=10$  & \ul{69.9} & 50.4 & 12.3 & \ul{45.8} & \ul{81.8} & 62.0 & 52.0 \\
$k=15$  & \ul{69.9} & 50.5 & \textbf{15.2} & \textbf{46.5} & 81.3 & \ul{62.1} & \textbf{52.7} \\
\bottomrule
\end{tabular}}
\caption{Per-benchmark accuracy for different hyperparameter settings ($k$) of Pattern-Aware Trailing Punctuation Filter on Korean data. The best scores are in bold, and the second-best scores are underlined.}
\label{tab:tuning-ko-patp-detail}
\end{table*}

After applying pattern-aware line-level deduplication with $r=50$ and $g=3$, we search over $k \in \{1, 3, 5, 7, 10, 15\}$. We find that $k=15$ yields the best average accuracy. Table \ref{tab:tuning-ko-patp-detail} reports the results. 

\section{Evaluation Method of Downstream Benchmarks}
\label{sec:app-benchmark-detail}

We utilize \texttt{lm-evaluation-harness}~\cite{eval-harness} and \texttt{vllm}~\cite{kwon2023vllm} for evaluation. Table ~\ref{tab:benchmark-statistics} summarizes the benchmarks we used and the number of examples in benchmarks.

\begin{table*}[t]
\centering
\begin{tabular}{@{}c|c|c|c|c@{}}
\toprule
Language                    & Benchmark         & Train     & Validation    & Test  \\ \midrule
\multirow{8}{*}{English}   & HellaSwag         & 39,905    & 10,042        & 10,003\\
                            & ARC Easy          & 2,251     & 570           & 2,376 \\
                            & CommonsenseQA     & 9,741     & 1,221         & 1,140  \\
                            & OpenBookQA        & 4,957     & 500           & 500    \\
                            & PIQA              & 16,113    & 1,838         & 3,084  \\
                            & SciQ              & 11,679    & 1,000         & 1,000  \\
                            & SIQA(Social IQa)  & 33,410    & 1,954         & N/A    \\
                            & SQuAD v1          & 87,599    & 10,570        & N/A    \\ \midrule
\multirow{5}{*}{Korean}    & KoBEST-HellaSwag  & 3,665     & 700           & 1,404    \\
                            & KoBEST-COPA       & 3,076     & 1,000         & 1,000  \\
                            & KorQuAD v1      & 60,407    & 5,774         & N/A    \\
                            & SNU Ko-ARC Easy    & -         & -             & 2,376  \\
                            & SNU Ko-LAMBADA     & -         & -             & 2,255  \\
\bottomrule
\end{tabular}
\caption{Number of examples in benchmarks. - denotes 'Not Applicable(N/A)'}
\label{tab:benchmark-statistics}

\end{table*}

\newcommand{\testsplit}[1]{We use the #1 split for evaluation.}

\newcommand{\mmlustyle}{We format questions and answers in MMLU-Style\cite{hendrycks2021mmlu}.}

\newcommand{\accnorm}{For each choice, we compute the negative log-likelihood of the ending tokens, normalized by the length of the ending. The choice with the highest normalized score is selected.}

\newcommand{\acc}{For each choice, we compute the negative log-likelihood of the ending tokens. The choice with the highest score is selected.}

\newcommand{\fone}{We compute the F1 score.}

\newcommand{\exactmatch}{We compute the exact match score.}

\newcommand{\instloose}{We compute instruction level loose accuracy.}

\newcommand{\passatone}{We report pass@1.}

\newcommand{\zeroshot}{We report the zero-shot accuracy for the task.}
\newcommand{\noexample}{We do not provide examples in the prompt.}

\newcommand{\fewshot}[1]{We report the #1-shot accuracy for the task.}

\newcommand{\fewshotsplit}[2]{}

\newcommand{\firstn}[2]{We select the first #2 examples from the #1 split.}

\newcommand{\randn}[2]{We randomly sample #2 examples from the #1 split.}

\newcommand{\avoidself}{If the test instance appears among the few-shot examples, we replace it by sampling an additional example.}

\newcommand{\greedydecoding}[0]{We use greedy decoding to sample the next token.}

\newcommand{\refertoprompt}[1]{Table~\ref{#1} describes the evaluation prompts.}


\subsection{English Downstream Tasks}

\subsubsection{HellaSwag~\cite{zellers2019hellaswag}}
\testsplit{validation}~\zeroshot ~\accnorm
~\refertoprompt{tab:eval-prompts-hellaswag}

\subsubsection{SQuAD V1.0~\cite{rajpurkar2016squad1}}
\testsplit{validation}~\fewshot{3}~\firstn{validation}{3}~\exactmatch ~\greedydecoding
~\refertoprompt{tab:eval-prompts-squadv1}

\subsubsection{CommonsenseQA~\cite{talmor2019commonsenseqa}}
\testsplit{validation}~\fewshot{5}~\firstn{validation}{5}~\accnorm
~\refertoprompt{tab:eval-prompts-commonsenseqa}

\subsubsection{OpenbookQA~\cite{mihaylov-etal-2018-suit}}
\testsplit{test}~\zeroshot ~\accnorm
~\refertoprompt{tab:eval-prompts-openbookqa}

\subsubsection{PIQA~\cite{bisk2020piqa}}
\testsplit{validation}~\zeroshot ~\accnorm
~\refertoprompt{tab:eval-prompts-piqa}

\subsubsection{SciQ~\cite{welbl2017sciencqa}}
\testsplit{test}~\zeroshot ~\accnorm
~\refertoprompt{tab:eval-prompts-sciq}

\subsubsection{Social IQa~\cite{sap2019socialiqa}}
\testsplit{validation}~\zeroshot ~\accnorm
~\refertoprompt{tab:eval-prompts-siqa}

\subsubsection{ARC-Easy~\cite{arc}}
\testsplit{test}~\fewshot{5}~\firstn{test}{5}~\accnorm
~\refertoprompt{tab:eval-prompts-arc}

\subsection{Korean Downstream Tasks}

\subsubsection{KoBEST-Hellaswag~\cite{jang-etal-2022-kobest}}
\testsplit{validation}~\zeroshot ~\accnorm
~\refertoprompt{tab:eval-prompts-kobest-hellaswag}

\subsubsection{KoBEST-COPA~\cite{jang-etal-2022-kobest}}
\testsplit{test}~\zeroshot ~\accnorm
~\refertoprompt{tab:eval-prompts-kobest-copa}

\subsubsection{KorQuAD V1.0~\cite{lim2019korquad1}}
\testsplit{validation}~\fewshot{5}~\firstn{validation}{5}~\exactmatch ~\greedydecoding
~\refertoprompt{tab:eval-prompts-korquadv1}

\subsubsection{SNU-Ko-ARC-Easy~\cite{kim2025thunder}}
\testsplit{test}~\zeroshot ~\accnorm
~\refertoprompt{tab:eval-prompts-snu-arc}

\subsubsection{SNU-Ko-LAMBADA~\cite{kim2025thunder}}
\testsplit{test}~\zeroshot ~\accnorm
~\refertoprompt{tab:eval-prompts-snu-lambada}

\section{Checklist for ARR Submission}

\label{sec:arr-checklist}
\paragraph{(A1) Limitations of the work}
Please refer to the Limitations section of the main text.

\paragraph{(A2) Potential risks of the work}

We utilize publicly-available data for the research in accordance with their license, so there are no copyright issues.

We do not disclose our processed data, so there would be no issues regarding violence, sexual abuse, and leak of personally identifiable information.


\paragraph{(B1) Citations of the used artifacts}
We tried our best to cite all papers, code repositories, and resources we used in the main content of the paper.


\paragraph{(B2) License or terms of use of the artifacts}

\paragraph{CommonCrawl Data.}
The CommonCrawl data is publicly released under CommonCrawl Terms of Use(ToU) which freely grants use of released data, including academic research. We follow the terms of use and restrict our purpose of data usage to academic research.


\paragraph{Benchmarks.}
We ensure all of the benchmarks are properly licensed for public use and appropriate for evaluating language models. CommonsenseQA and HellaSwag are licensed under MIT. OpenBookQA and PIQA licensed under Apache 2.0. SciQ is licensed under CC-BY-NC 3.0. Social IQa is licensed under CC-BY 4.0. ARC-Easy, KoBEST-COPA, KoBEST-HellaSwag, SNU Ko-ARC Easy, and SQuAD v1 are licensed under CC-BY-SA 4.0. KorQuAD v1.0 is licensed under CC-BY-ND 4.0. SNU Ko-LAMBADA is licensed under CC-BY-NC-SA 4.0.  


\paragraph{(B3) Proper use of existing artifacts and Intended use of created artifacts}

\paragraph{Commoncrawl}
We used text data derived from the Common Crawl corpus, which is made publicly available under its terms of use. The dataset's intended use is for academic and research purposes, and our use was fully consistent with these terms. We did not modify or redistribute the data. 

\paragraph{Benchmarks for evaluating this model.} 
For evaluating models, we utilize the test set of each benchmark. If it is impossible to utilize the test set for evaluation, we utilize the validation set. For more detail, see Appendix \ref{sec:app-benchmark-detail}.


\paragraph{(B4) Description of steps for removing personal identifiable information(PII) and offensive contents from data}

As our research scope is limited to the foremost stages of the data pipeline, we do not consider PII removal or safety filters in our research scope.
However, as we are not releasing our data nor trained models, there will be no safety or privacy risk.

\paragraph{(B5) Documentation of the artifacts}
We release our code for reproduction. See https://github.com/mcrl/pattern-aware-filtering.

\paragraph{(B6) Statistics for the data}

\paragraph{Data for pretraining}
See Appendix ~\ref{sec:app-detail-data} for detailed information on the pretraining data.
For statistics, see Table ~\ref{tab:filtered-data-detail}.

\paragraph{Benchmark} 
For benchmarks, see Table~\ref{tab:benchmark-statistics}

\paragraph{(C1) Descriptions of the number of parameters in the model, the total computational budget, and computing infrastructure}
See Appendix \ref{sec:app-experiment-detail} for the number of parameters in the model.
See Appendix \ref{sec:app-computation-train} for the total computational budget and computing infrastructure.

\paragraph{(C2) Details of experimental setup}
See Appendix \ref{sec:app-experiment-detail} for the experimental setup, and Appendix \ref{sec:tuning} for threshold tuning procedures.

\paragraph{(C3) Descriptive statistics about results}

We train each model twice and report the mean score measured from two different models. We disclose the full data in Appendix \ref{sec:repetition}.

\paragraph{(C4) Use of existing packages} 

\textbf{Data Extraction} We use \texttt{fasttext-langdetect} (v1.0.5)~\cite{joulin2016fasttext} for language identification in English, 
\texttt{numpy} (v1.2.6)~\cite{harris2020array} for building hashbins.

\textbf{Training} We use \texttt{torch} (v2.4.0)~\cite{paszke2019pytorch} and 
\texttt{transformers} (v4.51.1)~\cite{wolf2020transformers} for model implementation, 
\texttt{deepspeed} (v0.14.5)~\cite{rasley2020deepspeed} with ZeRO stage-1 for distributed training, 
and \texttt{flash-attn} (v2.5.8)~\cite{dao2022flashattention} for accelerating attention computation.

\textbf{Evaluation} We use \texttt{vllm} (v0.5.4)~\cite{kwon2023vllm} for faster inference and \texttt{lm-eval-harness} (v0.4.8)~\cite{eval-harness} for benchmark evaluation. 




\paragraph{(D1) Full text of instructions or disclaimers of any risks}
We do not use human annotators nor research with human participants.

\paragraph{(D2) Recruitment process and payment of paid participants}
We do not use human annotators nor research with human participants.

\paragraph{(D3) Consent from the used/curated data}
We do not use human annotators nor research with human participants.

\paragraph{(D4) Review of data collection protocol by an ethics review board}
We do not use human annotators nor research with human participants.

\paragraph{(D5) Basic demographic and geographic characteristics of the annotator population}
We do not use human annotators nor research with human participants.


\paragraph{(E1) Use of AI assistants}
We do not use AI assistants in our work.

\section{Experiment Results over Repetitions \label{sec:repetition}}

We disclose all measured accuracies over repeated experiments.

\begin{table*}[t]
\centering
\resizebox{\textwidth}{!}{%
\begin{tabular}{lc|cccccccc|cc}\toprule
Method &Seed &ARC(Easy) &CSQA &Hellaswag &OBQA &PIQA &SciQ &SIQA &SquAD(v1) &Mean(w.o SquAD) &Mean \\\midrule
Baseline &1200 &46.7 &36.9 &36.7 &31.8 &66.5 &64.4 &42.6 &6.0 &46.5 &41.4 \\
Baseline &5678 &46.7 &39.6 &37.1 &31.2 &66.3 &63.5 &42.6 &8.7 &46.7 &42.0 \\
LD &1200 &52.1 &44.5 &44.9 &35.0 &69.6 &73.2 &44.1 &0.1 &51.9 &45.4 \\
LD &5678 &51.4 &39.0 &45.6 &33.0 &70.1 &72.3 &43.2 &0.2 &50.7 &44.4 \\
LD+TF &1200 &45.9 &31.0 &47.9 &33.4 &70.9 &66.6 &43.2 &0.0 &48.4 &42.4 \\
LD+TF &5678 &40.8 &30.3 &47.8 &33.8 &71.1 &71.4 &42.9 &0.0 &48.3 &42.3 \\
PLD &1200 &53.1 &48.6 &46.0 &32.8 &70.6 &72.8 &43.5 &13.0 &52.5 &47.6 \\
PLD &5678 &54.9 &45.9 &45.4 &33.2 &71.4 &71.6 &43.4 &16.7 &52.3 &47.8 \\
PLD+PTF &1200 &55.6 &48.2 &47.5 &33.6 &71.3 &73.4 &44.4 &15.0 &53.4 &48.6 \\
PLD+PTF &5678 &54.0 &49.3 &48.0 &32.8 &71.5 &72.7 &42.4 &10.0 &53.0 &47.6 \\
\bottomrule
\end{tabular}}
\caption{Full Experiment Results for Table \ref{tab:exp-en-base}}
\label{tab:exp-en-base-full}
\end{table*}
\begin{table*}[t]
\centering
\resizebox{\textwidth}{!}{%
\begin{tabular}{lc|ccccc|cc}\toprule
KorQuAD(v1)
Method &Seed &KoBEST-COPA &KoBEST-Hellaswag &SNU-Ko-ARC(E) &SNU-Ko-Lambada &KorQuAD(v1) &Mean(w.o. KorQuAD) &Mean \\\midrule
Baseline &1200 &59.9 &42.4 &0.1 &37.4 &74.5 &53.5 &42.8 \\
Baseline &5678 &61.4 &41.8 &0.4 &38.0 &74.3 &53.9 &43.2 \\
LD &1200 &69.0 &50.8 &7.1 &45.8 &81.4 &61.7 &50.8 \\
LD &5678 &68.5 &49.8 &5.8 &46.0 &80.7 &61.2 &50.2 \\
LD+TF &1200 &69.8 &50.4 &0.1 &45.4 &82.0 &61.9 &49.5 \\
LD+TF &5678 &71.6 &51.0 &0.0 &46.1 &82.1 &62.7 &50.2 \\
PLD &1200 &70.3 &50.2 &3.7 &45.9 &81.6 &62.0 &50.3 \\
PLD &5678 &68.8 &51.2 &12.1 &46.5 &82.2 &62.2 &52.2 \\
PLD+PTF &1200 &71.5 &52.6 &4.3 &47.9 &80.0 &63.0 &51.3 \\
PLD+PTF &5678 &70.0 &53.4 &3.5 &48.9 &81.3 &63.4 &51.4 \\
\bottomrule
\end{tabular}}
\caption{Full Experiment Results for Table \ref{tab:exp-ko-base}}
\label{tab:exp-ko-base-full}
\end{table*}
\begin{table*}[t]
\centering
\resizebox{\textwidth}{!}{%
\begin{tabular}{lc|cccccccc|cc}\toprule
Method &Seed &ARC(Easy) &CSQA &Hellaswag &OBQA &PIQA &SciQ &SIQA &SquAD(v1) &Mean(w.o SquAD) &Mean \\\midrule
$r=50,g=1$ &1200 &50.9 &45.4 &42.0 &32.2 &68.3 &71.8 &43.0 &13.6 &50.5 &45.9 \\
$r=50,g=1$ &5678 &51.5 &45.9 &42.1 &31.8 &69.2 &70.3 &43.8 &14.7 &50.7 &46.2 \\
$r=50,g=3$ &1200 &49.7 &46.0 &42.0 &33.6 &68.2 &70.3 &43.3 &12.8 &50.4 &45.7 \\
$r=50,g=3$ &5678 &49.6 &42.8 &41.9 &32.4 &69.6 &72.1 &43.3 &15.3 &50.3 &45.9 \\
$r=50,g=5$ &1200 &50.1 &45.1 &42.5 &32.4 &68.5 &67.6 &43.3 &15.0 &49.9 &45.6 \\
$r=50,g=5$ &5678 &51.9 &43.2 &41.6 &30.2 &68.5 &70.5 &43.7 &14.7 &50.0 &45.6 \\
$r=50,g=10$ &1200 &50.2 &42.8 &41.7 &31.6 &67.2 &68.8 &43.2 &10.2 &49.4 &44.5 \\
$r=50,g=10$ &5678 &50.2 &43.6 &42.0 &32.8 &68.1 &69.4 &44.0 &8.5 &50.0 &44.8 \\
$r=50,g=15$ &1200 &50.2 &44.5 &41.5 &32.6 &68.1 &69.9 &43.9 &12.2 &50.1 &45.3 \\
$r=50,g=15$ &5678 &49.2 &43.2 &41.4 &32.0 &68.7 &70.8 &44.0 &12.5 &49.9 &45.2 \\
$r=50,g=20$ &1200 &49.8 &42.9 &41.9 &31.0 &68.1 &68.2 &43.2 &9.6 &49.3 &44.3 \\
$r=50,g=20$ &5678 &51.0 &43.7 &41.3 &32.4 &67.8 &69.5 &43.0 &7.8 &49.8 &44.6 \\
$r=100,g=1$ &1200 &50.8 &45.7 &41.9 &32.8 &68.7 &68.3 &43.0 &12.8 &50.2 &45.5 \\
$r=100,g=1$ &5678 &50.8 &43.2 &42.1 &32.8 &69.0 &69.3 &43.6 &12.4 &50.1 &45.4 \\
$r=100,g=3$ &1200 &51.7 &43.7 &41.9 &32.2 &67.8 &71.8 &42.8 &17.5 &50.3 &46.2 \\
$r=100,g=3$ &5678 &49.8 &42.6 &41.7 &34.8 &68.4 &67.0 &42.9 &13.2 &49.6 &45.1 \\
$r=100,g=5$ &1200 &49.7 &42.9 &42.4 &32.4 &69.5 &69.1 &43.3 &15.8 &49.9 &45.6 \\
$r=100,g=5$ &5678 &49.0 &42.8 &42.3 &30.2 &68.8 &70.2 &42.8 &10.7 &49.4 &44.6 \\
$r=100,g=10$ &1200 &50.8 &43.7 &42.4 &32.2 &69.1 &68.8 &43.5 &11.7 &50.1 &45.3 \\
$r=100,g=10$ &5678 &50.4 &43.7 &42.1 &32.2 &69.5 &67.0 &43.6 &11.3 &49.8 &45.0 \\
$r=100,g=15$ &1200 &49.2 &43.2 &42.1 &32.4 &68.7 &70.4 &43.1 &13.2 &49.9 &45.3 \\
$r=100,g=15$ &5678 &49.6 &43.6 &41.7 &33.2 &68.0 &69.9 &44.0 &8.6 &50.0 &44.8 \\
$r=100,g=20$ &1200 &48.8 &42.3 &41.3 &31.4 &69.3 &70.6 &43.1 &16.1 &49.5 &45.4 \\
$r=100,g=20$ &5678 &48.7 &41.8 &41.8 &34.2 &67.6 &68.3 &43.8 &9.7 &49.4 &44.5 \\
$r=500,g=1$ &1200 &51.4 &44.1 &42.2 &33.0 &68.8 &71.0 &43.2 &7.9 &50.5 &45.2 \\
$r=500,g=1$ &5678 &49.4 &43.2 &41.8 &32.0 &68.2 &69.2 &43.4 &15.4 &49.6 &45.3 \\
$r=500,g=3$ &1200 &49.5 &42.2 &41.9 &34.4 &68.6 &68.3 &43.3 &13.8 &49.7 &45.3 \\
$r=500,g=3$ &5678 &51.5 &44.4 &42.0 &33.8 &68.8 &70.3 &43.0 &13.4 &50.6 &45.9 \\
$r=500,g=5$ &1200 &49.8 &43.4 &41.6 &32.6 &68.5 &69.8 &43.6 &13.7 &49.9 &45.4 \\
$r=500,g=5$ &5678 &50.5 &43.3 &41.6 &31.8 &68.8 &72.6 &43.2 &15.8 &50.3 &45.9 \\
$r=1000,g=1$ &1200 &51.9 &43.9 &42.2 &32.6 &68.7 &70.9 &43.9 &16.1 &50.6 &46.3 \\
$r=1000,g=1$ &5678 &51.1 &44.6 &42.7 &32.6 &68.9 &69.7 &42.0 &16.0 &50.2 &45.9 \\
$r=1000,g=3$ &1200 &49.5 &42.9 &41.7 &32.0 &68.4 &70.6 &44.4 &15.5 &49.9 &45.6 \\
$r=1000,g=3$ &5678 &49.7 &44.6 &42.2 &34.0 &67.2 &70.0 &43.4 &14.4 &50.2 &45.7 \\
$r=1000,g=5$ &1200 &50.5 &45.0 &41.7 &31.2 &69.6 &71.8 &43.5 &17.9 &50.5 &46.4 \\
$r=1000,g=5$ &5678 &49.7 &41.9 &41.4 &32.0 &68.7 &68.9 &44.9 &18.1 &49.6 &45.7 \\
$r=1000,g=10$ &1200 &48.8 &43.7 &41.6 &32.4 &68.9 &68.2 &43.5 &12.8 &49.6 &45.0 \\
$r=1000,g=10$ &5678 &48.9 &41.9 &41.7 &33.0 &69.2 &69.0 &43.1 &14.7 &49.5 &45.2 \\
$r=1000,g=15$ &1200 &49.7 &43.4 &41.1 &32.8 &68.1 &69.5 &43.3 &11.6 &49.7 &44.9 \\
$r=1000,g=15$ &5678 &49.2 &43.4 &41.6 &32.2 &67.7 &69.0 &42.8 &9.1 &49.4 &44.4 \\
$r=1000,g=20$ &1200 &51.1 &42.5 &41.4 &31.2 &68.0 &68.7 &43.3 &10.4 &49.5 &44.6 \\
$r=1000,g=20$ &5678 &49.7 &42.8 &41.0 &33.4 &67.5 &67.6 &43.0 &6.4 &49.3 &43.9 \\
$r=2500,g=1$ &1200 &50.1 &45.9 &42.0 &32.8 &68.1 &71.1 &42.8 &17.6 &50.4 &46.3 \\
$r=2500,g=1$ &5678 &50.9 &45.5 &42.5 &32.0 &68.6 &68.8 &43.1 &10.5 &50.2 &45.2 \\
$r=2500,g=3$ &1200 &51.7 &45.9 &41.5 &35.4 &68.7 &68.1 &43.1 &4.1 &50.6 &44.8 \\
$r=2500,g=3$ &5678 &50.8 &43.6 &41.6 &33.0 &68.9 &69.8 &43.7 &14.6 &50.2 &45.7 \\
$r=2500,g=5$ &1200 &50.5 &43.8 &42.1 &32.4 &68.3 &71.7 &44.2 &12.8 &50.4 &45.7 \\
$r=2500,g=5$ &5678 &49.7 &45.0 &41.9 &31.6 &68.1 &70.7 &43.2 &13.8 &50.0 &45.5 \\
$r=5000,g=1$ &1200 &52.5 &44.0 &42.3 &32.0 &69.1 &70.5 &43.6 &16.7 &50.6 &46.3 \\
$r=5000,g=1$ &5678 &50.2 &42.2 &41.8 &32.0 &68.5 &70.9 &42.8 &17.3 &49.8 &45.7 \\
$r=5000,g=3$ &1200 &50.4 &42.8 &41.7 &31.8 &68.3 &68.8 &43.3 &11.2 &49.6 &44.8 \\
$r=5000,g=3$ &5678 &50.6 &43.0 &42.1 &32.8 &67.9 &71.2 &42.5 &6.9 &50.0 &44.6 \\
$r=5000,g=5$ &1200 &49.7 &42.9 &41.3 &31.8 &69.2 &68.9 &43.8 &3.3 &49.7 &43.9 \\
$r=5000,g=5$ &5678 &50.1 &45.0 &41.5 &32.0 &68.1 &70.6 &42.5 &11.8 &50.0 &45.2 \\
\bottomrule
\end{tabular}}
\caption{Full Experiment Results for Table \ref{tab:tuning-en-pald}}
\label{tab:tuning-en-pald-full}
\end{table*}
\begin{table*}[t]
\centering
\resizebox{\textwidth}{!}{%
\begin{tabular}{lc|ccccc|cc}\toprule
Method &Seed &KoBEST-COPA &KoBEST-Hellaswag &KorQuAD(v1) &SNU-Ko-ARC(E) &SNU-Ko-Lambada &Mean(w.o. KorQuAD) &Mean \\\midrule
$r=50,g=1$ &1200 &68.8 &51.0 &14.7 &44.2 &81.3 &61.3 &52.0 \\
$r=50,g=1$ &5678 &70.0 &50.6 &6.7 &46.3 &81.5 &62.1 &51.0 \\
$r=50,g=3$ &1200 &68.2 &51.4 &22.0 &43.3 &81.0 &61.0 &53.2 \\
$r=50,g=3$ &5678 &67.9 &50.6 &14.4 &45.2 &80.4 &61.0 &51.7 \\
$r=50,g=5$ &1200 &67.6 &49.8 &7.4 &45.2 &80.6 &60.8 &50.1 \\
$r=50,g=5$ &5678 &67.7 &47.4 &12.8 &43.6 &79.8 &59.6 &50.2 \\
$r=100,g=1$ &1200 &67.0 &52.0 &13.8 &43.9 &81.6 &61.1 &51.6 \\
$r=100,g=1$ &5678 &67.3 &49.2 &11.2 &45.3 &81.9 &60.9 &51.0 \\
$r=100,g=3$ &1200 &68.6 &48.4 &17.3 &43.9 &80.4 &60.3 &51.7 \\
$r=100,g=3$ &5678 &68.0 &50.8 &10.2 &43.1 &80.7 &60.6 &50.6 \\
$r=100,g=5$ &1200 &68.2 &50.2 &16.6 &44.1 &81.2 &60.9 &52.1 \\
$r=100,g=5$ &5678 &67.3 &50.0 &18.6 &44.9 &81.9 &61.0 &52.6 \\
$r=500,g=1$ &1200 &69.1 &50.2 &11.7 &44.9 &80.2 &61.1 &51.2 \\
$r=500,g=1$ &5678 &67.7 &50.0 &3.5 &43.0 &81.2 &60.5 &49.1 \\
$r=500,g=3$ &1200 &69.0 &50.0 &17.5 &43.6 &80.7 &60.8 &52.2 \\
$r=500,g=3$ &5678 &66.3 &50.0 &17.2 &45.7 &80.8 &60.7 &52.0 \\
$r=500,g=5$ &1200 &67.1 &50.2 &15.9 &43.4 &80.2 &60.2 &51.4 \\
$r=500,g=5$ &5678 &67.2 &50.8 &12.5 &43.7 &80.5 &60.6 &51.0 \\
$r=1000,g=1$ &1200 &69.4 &47.0 &15.2 &44.7 &80.6 &60.4 &51.4 \\
$r=1000,g=1$ &5678 &66.8 &52.4 &10.6 &43.9 &80.7 &61.0 &50.9 \\
$r=1000,g=3$ &1200 &68.4 &48.6 &11.9 &44.0 &79.7 &60.2 &50.5 \\
$r=1000,g=3$ &5678 &67.2 &48.6 &8.4 &44.5 &80.5 &60.2 &49.8 \\
$r=1000,g=5$ &1200 &66.3 &48.8 &12.9 &43.6 &81.6 &60.1 &50.6 \\
$r=1000,g=5$ &5678 &68.1 &49.2 &12.6 &43.1 &80.8 &60.3 &50.8 \\
$r=2500,g=1$ &1200 &67.8 &47.0 &17.5 &45.7 &79.9 &60.1 &51.6 \\
$r=2500,g=1$ &5678 &68.9 &48.8 &15.7 &43.6 &80.1 &60.3 &51.4 \\
$r=2500,g=3$ &1200 &68.8 &49.6 &12.4 &44.9 &80.5 &60.9 &51.2 \\
$r=2500,g=3$ &5678 &68.8 &49.8 &12.0 &43.9 &80.5 &60.8 &51.0 \\
$r=2500,g=5$ &1200 &66.4 &50.0 &17.2 &44.1 &80.4 &60.2 &51.6 \\
$r=2500,g=5$ &5678 &67.1 &48.6 &13.4 &44.4 &80.6 &60.2 &50.8 \\
$r=5000,g=1$ &1200 &68.0 &48.4 &17.8 &44.6 &80.5 &60.4 &51.9 \\
$r=5000,g=1$ &5678 &68.6 &49.8 &4.8 &44.4 &80.9 &60.9 &49.7 \\
$r=5000,g=3$ &1200 &66.6 &47.8 &8.5 &44.7 &79.7 &59.7 &49.5 \\
$r=5000,g=3$ &5678 &67.9 &49.8 &13.2 &43.9 &80.5 &60.5 &51.1 \\
$r=5000,g=5$ &1200 &67.6 &50.2 &8.4 &43.5 &80.1 &60.4 &50.0 \\
$r=5000,g=5$ &5678 &67.0 &50.8 &11.1 &43.6 &80.0 &60.3 &50.5 \\
\bottomrule
\end{tabular}}
\caption{Full Experiment Results for Table \ref{tab:tuning-ko-pald}}
\label{tab:tuning-ko-pald-full}
\end{table*}
\begin{table*}[t]
\centering
\resizebox{\textwidth}{!}{%
\begin{tabular}{lc|cccccccc|cc}\toprule
Method &Seed &ARC(Easy) &CSQA &Hellaswag &OBQA &PIQA &SciQ &SIQA &SquAD(v1) &Mean(w.o SquAD) &Mean \\\midrule
$k=1$ &1200 &50.8 &41.9 &43.8 &33.6 &69.9 &67.5 &42.6 &0.8 &50.0 &43.8 \\
$k=1$ &5678 &49.4 &44.4 &43.9 &33.0 &70.7 &70.6 &44.1 &0.8 &50.9 &44.6 \\
$k=3$ &1200 &51.7 &45.2 &44.1 &34.4 &69.6 &68.7 &43.1 &14.2 &51.0 &46.4 \\
$k=3$ &5678 &52.5 &47.1 &44.1 &33.6 &69.9 &73.5 &42.7 &7.5 &51.9 &46.4 \\
$k=5$ &1200 &52.7 &43.7 &43.3 &32.2 &70.3 &69.9 &43.5 &7.1 &50.8 &45.3 \\
$k=5$ &5678 &51.3 &45.6 &43.9 &31.6 &69.5 &67.9 &42.6 &8.4 &50.4 &45.1 \\
$k=7$ &1200 &51.3 &46.2 &43.3 &33.0 &70.2 &70.0 &43.3 &10.7 &51.0 &46.0 \\
$k=7$ &5678 &51.6 &42.6 &43.4 &31.6 &69.2 &71.1 &43.7 &14.2 &50.4 &45.9 \\
$k=10$ &1200 &50.5 &46.6 &43.5 &33.6 &70.1 &69.8 &42.8 &16.4 &51.0 &46.7 \\
$k=10$ &5678 &53.1 &44.3 &43.6 &32.8 &68.8 &71.3 &43.9 &8.7 &51.1 &45.8 \\
$k=15$ &1200 &52.5 &43.8 &43.4 &31.2 &68.7 &71.4 &43.2 &15.0 &50.6 &46.2 \\
$k=15$ &5678 &51.6 &44.3 &43.2 &32.6 &69.8 &69.5 &43.9 &13.4 &50.7 &46.0 \\
\bottomrule
\end{tabular}}
\caption{Full Experiment Results for Table \ref{tab:tuning-en-patp-detail}}
\label{tab:tuning-en-patp-full}
\end{table*}
\begin{table*}[t]
\centering
\resizebox{\textwidth}{!}{%
\begin{tabular}{lc|ccccc|cc}\toprule
Method &Seed &KoBEST-COPA &KoBEST-Hellaswag &KorQuAD(v1) &SNU-Ko-ARC(E) &SNU-Ko-Lambada &Mean(w.o. KorQuAD) &Mean \\\midrule
$k=1$ &1200 &68.2 &51.4 &12.6 &44.0 &81.4 &61.2 &51.5 \\
$k=1$ &5678 &69.7 &49.6 &8.8 &45.9 &81.2 &61.6 &51.0 \\
$k=3$ &1200 &71.2 &50.8 &10.5 &46.3 &81.2 &62.4 &52.0 \\
$k=3$ &5678 &69.3 &50.8 &18.8 &44.1 &80.8 &61.3 &52.8 \\
$k=5$ &1200 &70.0 &52.2 &12.2 &45.6 &80.9 &62.2 &52.2 \\
$k=5$ &5678 &69.1 &52.0 &15.5 &45.8 &81.8 &62.2 &52.8 \\
$k=7$ &1200 &69.8 &48.8 &10.3 &45.4 &82.1 &61.5 &51.3 \\
$k=7$ &5678 &70.0 &51.2 &10.3 &44.4 &81.9 &61.9 &51.6 \\
$k=10$ &1200 &69.9 &49.6 &11.1 &45.3 &81.3 &61.5 &51.5 \\
$k=10$ &5678 &70.0 &51.2 &13.4 &46.3 &82.3 &62.5 &52.6 \\
$k=15$ &1200 &69.2 &51.2 &15.0 &45.6 &81.3 &61.8 &52.5 \\
$k=15$ &5678 &70.7 &49.8 &15.3 &47.5 &81.3 &62.3 &52.9 \\
\bottomrule
\end{tabular}}
\caption{Full Experiment Results for Table \ref{tab:tuning-ko-patp-detail}}
\label{tab:tuning-ko-patp-full}
\end{table*}


\begin{table*}[htbp]
\centering
\begin{tabular}{@{}p{0.10\textwidth}|p{0.90\textwidth}@{}}
\toprule

\textbf{Problem} & \tablecell{
\textbf{Context:} \\
A cat is sitting in a cat bed. It is licking its paw. it \\
\\
\textbf{Choices:} \\
- then wipes its paw on its ear. \\
- crawls to the door. \\
- 's laying down about six inches away from the camera. \\
- crawls on one foot to get closer to the camera. \\

\\
\textbf{Answer:} \\
then wipes its paw on its ear..
}
\\ \midrule
 
\textbf{Context} & \tablecell{
A cat is sitting in a cat bed. It is licking its paw. it
}    
\\ \midrule
 
\textbf{Endings} & \tablecell{
then wipes its paw on its ear.
}
\\ \bottomrule
 
\end{tabular}%
\caption{Evaluation prompts for Hellaswag}
\label{tab:eval-prompts-hellaswag}

\end{table*}
\begin{table*}[htbp]
\centering
\begin{tabular}{@{}p{0.10\textwidth}|p{0.90\textwidth}@{}}
\toprule

\textbf{Problem} & \tablecell{
\textbf{Title:} \\
Super\_Bowl\_50 \\
\\
\textbf{Background:} \\
Super Bowl 50 was an American football game to determine the champion of the National Football League (NFL) for the 2015 season. 
The American Football Conference (AFC) champion Denver Broncos defeated the National Football Conference (NFC) champion Carolina Panthers 24–10 to earn their third Super Bowl title. 
The game was played on February 7, 2016, at Levi's Stadium in the San Francisco Bay Area at Santa Clara, California. 
As this was the 50th Super Bowl, the league emphasized the "golden anniversary" with various gold-themed initiatives, as well as temporarily suspending the tradition of naming each Super Bowl game with Roman numerals (under which the game would have been known as "Super Bowl L"), so that the logo could prominently feature the Arabic numerals 50. 
(\textit{content abbreviated}) \\
\\
\textbf{Question:} \\
Which NFL team represented the AFC at Super Bowl 50? \\
\\
\textbf{Answer:} \\
Denver Broncos
}
\\ \midrule
\textbf{Prompt} & \tablecell{
Read the passage and answer the question based on the context provided. \\\\
Title: University\_of\_Notre\_Dame \\\\
Background: Architecturally, the school has a Catholic character... 
It is a replica of the grotto at Lourdes, France where the Virgin Mary reputedly appeared to Saint Bernadette Soubirous in 1858. (\textit{content abbreviated})
 \\\\
Q: To whom did the Virgin Mary allegedly appear in 1858 in Lourdes France? \\\\
A: Saint Bernadette Soubirous \\\\
Title: Super\_Bowl\_50 \\\\
Background: Super Bowl 50 was an American football game to determine the champion of the National Football League (NFL) for the 2015 season... (\textit{content abbreviated}) \\\\
Q: Which NFL team represented the AFC at Super Bowl 50? \\\\
A:
}
\\ \midrule

\tablecell{\textbf{Expected} \\ \textbf{Output}} & \tablecell{
Denver Broncos
}
\\ \bottomrule

\end{tabular}%
\caption{Evaluation prompts for SQuAD v1}
\label{tab:eval-prompts-squadv1}
\end{table*}


 
 
 

\begin{table*}[htbp]
\centering
\begin{tabular}{@{}p{0.10\textwidth}|p{0.90\textwidth}@{}}
\toprule

\textbf{Problem} & \tablecell{
\textbf{Title:} \\
Revolving\_Door \\
\\
\textbf{Question:} \\
A revolving door is convenient for two direction travel, but it also serves as a security measure at a what? \\
\\
\textbf{Choices:} \\
- bank \\
- library \\
- department store \\ 
- mall \\
- new york \\
\\
\textbf{Answer:} \\
bank
}
\\ \midrule

\textbf{Context} & \tablecell{
Question: The sanctions against the school were a punishing blow, and they seemed to what the efforts the school had made to change? \\
Answer: ignore \\\\

Question: Sammy wanted to go to where the people were. Where might he go? \\
Answer: populated areas \\\\

Question: To locate a choker not located in a jewelry box or boutique where would you go? \\
Answer: jewelry store \\\\

Question: Google Maps and other highway and street GPS services have replaced what? \\
Answer: atlas \\\\

Question: The fox walked from the city into the forest, what was it looking for? \\
Answer: natural habitat \\\\

Question: A revolving door is convenient for two direction travel, but it also serves as a security measure at a what? \\\\
Answer:
}
\\ \midrule

\tablecell{\textbf{Endings}} & \tablecell{
bank
}
\\ \bottomrule
 

\end{tabular}%
\caption{Evaluation prompts for CommonsenseQA}
\label{tab:eval-prompts-commonsenseqa}

\end{table*}

\begin{table*}[htbp]
\centering
\begin{tabular}{@{}p{0.10\textwidth}|p{0.90\textwidth}@{}}
\toprule

\textbf{Problem} & \tablecell{
\textbf{Context:} \\
A person wants to start saving money so that they can afford a nice vacation at the end of the year. After looking over their budget and expenses, they decide the best way to save money is to \\
\\
\textbf{Choices:} \\
- make more phone calls\\
- quit eating lunch out\\
- buy less with monopoly money\\
- have lunch with friends\\
\\
\textbf{Answer:} \\
quit eating lunch out
}
\\ \midrule
 
\textbf{Context} & \tablecell{
A person wants to start saving money so that they can afford a nice vacation at the end of the year. After looking over their budget and expenses, they decide the best way to save money is to
}    
\\ \midrule
 
\textbf{Endings} & \tablecell{
quit eating lunch out
}
\\ \bottomrule
 
\end{tabular}%
\caption{Evaluation prompts for OpenbookQA}
\label{tab:eval-prompts-openbookqa}

\end{table*}
\begin{table*}[htbp]
\centering
\begin{tabular}{@{}p{0.10\textwidth}|p{0.90\textwidth}@{}}
\toprule

\textbf{Problem} & \tablecell{
\textbf{Question:} \\
How do I ready a guinea pig cage for its new occupants? \\
\\
\textbf{Choices:} \\
- Provide the guinea pig with a cage full of a few inches of bedding made of ripped paper strips, you will also need to supply it with a water bottle and a food dish. \\
- Provide the guinea pig with a cage full of a few inches of bedding made of ripped jeans material, you will also need to supply it with a water bottle and a food dish. \\
\\
\textbf{Answer:} \\
Provide the guinea pig with a cage full of a few inches of bedding made of ripped paper strips, you will also need to supply it with a water bottle and a food dish.
}
\\ \midrule

\textbf{Context} & \tablecell{
How do I ready a guinea pig cage for its new occupants? 
}
\\ \midrule

\tablecell{\textbf{Endings}} & \tablecell{
Provide the guinea pig with a cage full of a few inches of bedding made of ripped paper strips, you will also need to supply it with a water bottle and a food dish.
}
\\ \bottomrule

\end{tabular}%
\caption{Evaluation prompts for PIQA}
\label{tab:eval-prompts-piqa}
\end{table*}

\begin{table*}[htbp]
\centering
\begin{tabular}{@{}p{0.10\textwidth}|p{0.90\textwidth}@{}}
\toprule

\textbf{Problem} & \tablecell{
\textbf{Question:} \\
Compounds that are capable of accepting electrons, such as O\textsubscript{2} or F\textsubscript{2}, are called what? \\
\\
\textbf{Choices:} \\
- antioxidants \\
- Oxygen \\
- residues \\
- oxidants \\
\\
\textbf{Answer:} \\
oxidants
}
\\ \midrule

\textbf{Context} & \tablecell{
Oxidants and Reductants. Compounds that are capable of accepting electrons, such as O\textsubscript{2} or F\textsubscript{2}, are called oxidants (or oxidizing agents) because they can oxidize other compounds. In the process of accepting electrons, an oxidant is reduced. Compounds that are capable of donating electrons, such as sodium metal or cyclohexane (C\textsubscript{6}H\textsubscript{12}), are called reductants (or reducing agents) because they can cause the reduction of another compound. In the process of donating electrons, a reductant is oxidized. 
}
\\ \midrule

\tablecell{\textbf{Endings}} & \tablecell{
oxidants
}
\\ \bottomrule

\end{tabular}%
\caption{Evaluation prompts for SciQ}
\label{tab:eval-prompts-sciq}
\end{table*}

\begin{table*}[htbp]
\centering
\begin{tabular}{@{}p{0.10\textwidth}|p{0.90\textwidth}@{}}
\toprule

\textbf{Problem} & \tablecell{
\textbf{Question:} \\
What does Tracy need to do before this? \\
\\
\textbf{Choices:} \\
- make a new plan \\
- Go home and see Riley \\
- Find somewhere to go \\
\\
\textbf{Answer:} \\
Find somewhere to go
}
\\ \midrule

\textbf{Context} & \tablecell{
Tracy didn't go home that evening and resisted Riley's attacks.
}
\\ \midrule

\tablecell{\textbf{Endings}} & \tablecell{
Find somewhere to go
}
\\ \bottomrule

\end{tabular}%
\caption{Evaluation prompts for SIQA}
\label{tab:eval-prompts-siqa}
\end{table*}

\begin{table*}[htbp]
\centering
\begin{tabular}{@{}p{0.10\textwidth}|p{0.90\textwidth}@{}}
\toprule

\textbf{Problem} & \tablecell{
\textbf{Question:} \\
Which statement best explains why photosynthesis is the foundation of most food webs? \\
\\
\textbf{Choices:} \\
- Sunlight is the source of energy for nearly all ecosystems. \\
- Most ecosystems are found on land instead of in water. \\
- Carbon dioxide is more available than other gases. \\
- The producers in all ecosystems are plants. \\
\\
\textbf{Answer:} \\
Sunlight is the source of energy for nearly all ecosystems.
}
\\ \midrule

\textbf{Context} & \tablecell{
Question: Which factor will most likely cause a person to develop a fever? \\
Answer: a bacterial population in the bloodstream \\\\
Question: Lichens are symbiotic organisms made of green algae and fungi. What do the green algae supply to the fungi in this symbiotic relationship? \\
Answer: food \\\\
Question: When a switch is used in an electrical circuit, the switch can \\
Answer: stop and start the flow of current. \\\\
Question: Which of the following is an example of an assistive device? \\
Answer: contact lens \\\\
Question: Rocks are classified as igneous, metamorphic, or sedimentary according to \\
Answer: how they formed \\\\
Question: Which statement best explains why photosynthesis is the foundation of most food webs? \\\\
Answer:
}
\\ \midrule

\textbf{Endings} & \tablecell{
Sunlight is the source of energy for nearly all ecosystems.
}
\\ \bottomrule

\end{tabular}%
\caption{Evaluation prompts for ARC-Easy}
\label{tab:eval-prompts-arc}
\end{table*}

\begin{table*}[htbp]
\centering
\begin{tabular}{@{}p{0.10\textwidth}|p{0.45\textwidth}|p{0.45\textwidth}@{}}
\toprule
 & Content in Korean & Content in English \\ \midrule

\textbf{Problem} & \tablecell{
\textbf{Context:} \\
엄마는 외출한 아들에게 메모지를 주며 장을 봐올 것을 부탁한다. 아들은 엄마의 심부름을 하기 위해 마트에 간다. 아들은 마트에서 카트를 챙기고 엄마가 준 메모지를 확인한다. \\
\\
\textbf{Choices:} \\
- 메모지를 확인하며 메모에 적힌 품목을 카트에 넣는다. \\
- 물건이 담긴 카트를 계산대에 가져간다. \\
- 아들이 계산이 완료된 물건을 장바구니에 담는다. \\
- 주인이 물건의 바코드를 찍고 카드를 받아 계산한다. \\
\\
\textbf{Answer:} \\
메모지를 확인하며 메모에 적힌 품목을 카트에 넣는다.
} & \tablecell{
\textbf{Context:} \\
A mother gives her son, who is going out, a note and asks him to do the grocery shopping. The son goes to the supermarket to run his mother’s errand. At the store, he grabs a cart and checks the note his mother gave him. \\
\\
\textbf{Choices:} \\
- He checks the note and puts the listed items into the cart. \\
- He brings the filled cart to the checkout counter. \\
- The son puts the purchased items into a shopping bag. \\
- The cashier scans the barcodes and takes the card for payment. \\
\\
\textbf{Answer:} \\
He checks the note and puts the listed items into the cart.
}
\\ \midrule

\textbf{Context} & \tablecell{
엄마는 외출한 아들에게 메모지를 주며 장을 봐올 것을 부탁한다. 아들은 엄마의 심부름을 하기 위해 마트에 간다. 아들은 마트에서 카트를 챙기고 엄마가 준 메모지를 확인한다.
} & \tablecell{
A mother gives her son, who is going out, a note and asks him to do the grocery shopping. The son goes to the supermarket to run his mother’s errand. At the store, he grabs a cart and checks the note his mother gave him.
}
\\ \midrule

\textbf{Endings} & \tablecell{
메모지를 확인하며 메모에 적힌 품목을 카트에 넣는다.
} & \tablecell{
He checks the note and puts the listed items into the cart.
}
\\ \bottomrule

\end{tabular}%
\caption{Evaluation prompts for KoBEST-Hellaswag}
\label{tab:eval-prompts-kobest-hellaswag}
\end{table*}

\begin{table*}[htbp]
\centering
\begin{tabular}{@{}p{0.10\textwidth}|p{0.45\textwidth}|p{0.45\textwidth}@{}}
\toprule
 & Content in Korean & Content in English \\ \midrule

\textbf{Problem} & \tablecell{
\textbf{Premise:} \\
새로 산 바지의 허리가 컸다. \\
\\
\textbf{Question:} \\
결과 \\
\\
\textbf{Alternatives:} \\
- 바지의 밑단을 잘랐다. \\
- 벨트를 바지에 끼워 사이즈를 조절했다. \\
\\
\textbf{Answer:} \\
2
} & \tablecell{
\textbf{Premise:} \\
The waist of the newly purchased pants was too large. \\
\\
\textbf{Question:} \\
Result \\
\\
\textbf{Alternatives:} \\
- The hem of the pants was cut. \\
- A belt was used to adjust the size of the pants. \\
\\
\textbf{Answer:} \\
2
}
\\ \midrule

\textbf{Context} & \tablecell{
새로 산 바지의 허리가 컸다. 그래서
} & \tablecell{
The waist of the newly purchased pants was too large. So
}
\\ \midrule

\textbf{Endings} & \tablecell{
바지의 밑단을 잘랐다.
} & \tablecell{
The hem of the pants was cut.
}
\\ \bottomrule

\end{tabular}%
\caption{Evaluation prompts for KoBEST-COPA}
\label{tab:eval-prompts-kobest-copa}
\end{table*}
 
\newcommand{\dbnewlinechar}{\textbackslash n\textbackslash n}

\begin{table*}[htbp]
\centering
\begin{tabular}{@{}p{0.10\textwidth}|p{0.45\textwidth}|p{0.45\textwidth}@{}}
\toprule
 & Content in Korean & Content in English \\ 
\midrule

\textbf{Problem} & \tablecell{
\textbf{Title:} 알렉산더\_헤이그 \\
\\
\textbf{Context:} \\
알렉산더 메이그스 헤이그 2세(영어: Alexander Meigs Haig, Jr., 1924년 12월 2일 \textasciitilde 2010년 2월 20일)는 미국의 국무 장관을 지낸 미국의 군인, 관료 및 정치인이다. 로널드 레이건 대통령 밑에서 국무장관을 지냈으며, \textit{(후략)} \\
\\
\textbf{Question:} \\
로널드 레이건 대통령 밑에서 일한 국무장관은 누구인가? \\
\\
\textbf{Answer:} \\
알렉산더 메이그스 헤이그 2세
} & \tablecell{
\textbf{Title:} Alexander\_Haig \\
\\
\textbf{Context:} \\
Alexander Meigs Haig Jr. (December 2, 1924 \textasciitilde February 20, 2010) was an American military officer, bureaucrat, and politician who served as the U.S. Secretary of State. He served as Secretary of State under President Ronald Reagan. \textit{(truncated)} \\
\\
\textbf{Question:} \\
Who served as Secretary of State under President Ronald Reagan? \\
\\
\textbf{Answer:} \\
Alexander Meigs Haig Jr.
}
\\ \midrule

\textbf{Context} & \tablecell{
주어진 배경지식을 읽고, 문제에 대한 답을 지문에서 찾아 답변하세요. \\\\
주제: 파우스트\_서곡 \\\\
배경지식: 1839년 바그너는 괴테의 파우스트를 처음 읽고 그 내용에 마음이 끌려 이를 소재로 해서 하나의 교향곡을 쓰려는 ... \textit{(후략)} \\\\
질문: 바그너는 괴테의 파우스트를 읽고 무엇을 쓰고자 했는가? \\\\
답: 교향곡 \\\\
\textit{(위와 같이 주어진 배경지식과 질문에 대한 답변이 주어지며, 각 질문과 답변은 한 줄의 빈 줄로 구분된다.)} \\\\
주제: 알렉산더\_헤이그 \\\\
배경지식: 알렉산더 메이그스 헤이그 2세(영어: Alexander Meigs Haig, Jr., 1924년 12월 2일 \textasciitilde 2010년 2월 20일)는 미국의 국무 장관을 지낸 미국의 군인, 관료 및 정치인이다. 로널드 레이건 대통령 밑에서 국무장관을 지냈으며, \textit{(후략)} \\\\
질문: 로널드 레이건 대통령 밑에서 일한 국무장관은 누구인가? \\\\
답:
} & \tablecell{
Read the given background knowledge and find the answer to the question from the passage. \\\\
Topic: Faust\_Overture \\\\
Background: In 1839, Wagner read Goethe’s *Faust* for the first time and was inspired by its content, leading him to compose a symphony based on it... \textit{(truncated)} \\\\
Question: After reading Goethe’s *Faust*, what did Wagner intend to compose? \\\\
Answer: Symphony \\\\
\textit{(As shown above, each question–answer pair is provided, separated by a blank line.)} \\\\
Topic: Alexander\_Haig \\\\
Background: Alexander Meigs Haig Jr. (December 2, 1924 \textasciitilde February 20, 2010) was an American military officer, bureaucrat, and politician who served as the U.S. Secretary of State. He served under President Ronald Reagan. \textit{(truncated)} \\\\
Question: Who served as Secretary of State under President Ronald Reagan? \\\\
Answer:
}
\\ \midrule

\textbf{Endings} & \tablecell{
알렉산더 메이그스 헤이그 2세
} & \tablecell{
Alexander Meigs Haig Jr.
}
\\ \bottomrule

\end{tabular}%
\caption{Evaluation prompts for KorQuAD (v1)}
\label{tab:eval-prompts-korquadv1}
\end{table*}

\begin{table*}[htbp]
\centering
\begin{tabular}{@{}p{0.10\textwidth}|p{0.45\textwidth}|p{0.45\textwidth}@{}}
\toprule
 & Content in Korean & Content in English \\ \midrule

\textbf{Problem} & \tablecell{
\textbf{Question:} \\
곰팡이 포자가 호흡기로 들어가는 것을 막기 위해 사용되는 안전 장비에는 어떤 것이 있나요? \\
\\
\textbf{Choices:} \\
- 보안경 \\
- 호흡 마스크 \\
- 고무 장갑 \\
- 납 앞치마 \\
\\
\textbf{Answer:} \\
호흡 마스크
} & \tablecell{
\textbf{Question:} \\
What safety equipment is used to prevent mold spores from entering the respiratory system? \\
\\
\textbf{Choices:} \\
- Safety goggles \\
- Respiratory mask \\
- Rubber gloves \\
- Lead apron \\
\\
\textbf{Answer:} \\
Respiratory mask
}
\\ \midrule

\textbf{Context} & \tablecell{
질문: 따오기는 한국의 습지에서 서식하던 조류로, 현재 멸종 위기에 처해 있습니다. 다음 중 따오기 멸종의 원인으로 가장 가능성이 높은 것은 무엇인가요? \\
답변: 과도한 사냥 \\\\
\textit{(위와 같이 질문과 답변이 주어지며, 각 질문과 답변은 한 줄의 빈 줄로 구분된다.)} \\\\
질문: 곰팡이 포자가 호흡기로 들어가는 것을 막기 위해 사용되는 안전 장비에는 어떤 것이 있나요? \\
답변:
} & \tablecell{
Question: The crested ibis, once a bird inhabiting wetlands in Korea, is now facing extinction. Which of the following is the most likely cause of its extinction? \\
Answer: Overhunting \\\\
\textit{(As shown above, each question and answer pair is given, separated by a single blank line.)} \\\\
Question: What safety equipment is used to prevent mold spores from entering the respiratory system? \\
Answer:
}
\\ \midrule

\textbf{Endings} & \tablecell{
호흡 마스크
} & \tablecell{
Respiratory mask
}
\\ \bottomrule

\end{tabular}%
\caption{Evaluation prompts for SNU-Ko-ARC-Easy}
\label{tab:eval-prompts-snu-arc}
\end{table*}

\begin{table*}[htbp]
\centering
\begin{tabular}{@{}p{0.10\textwidth}|p{0.45\textwidth}|p{0.45\textwidth}@{}}
\toprule
 & Content in Korean & Content in English \\ 
\midrule

\textbf{Problem} & \tablecell{
\textbf{Context:} \\
전차는 또 한 대 지나갔다. 승강대에 빈틈이 조금 있을 뿐, 미리 올라타서 가운데 숨어 있기에는 가장 적절한 전차였었다. 나는 혀를 한번 차고 담배를 꺼내어 붙여 물었다. 그 전차는 ○의 아내의 타는 정류장 앞에서 잠깐 멎었다가 다시 떠났다. 그러나 한 간을 나아가지 않아서 그 전차는 다시 멎었다. 나는 무심히 먹기 시작한 \_를 내어던지고 그편을 향하여 돌아섰다. \\
\\
\textbf{Choices:} \\
- 담배 \\
- 아내 \\
\\
\textbf{Answer:} \\
담배
} & \tablecell{
\textbf{Context:} \\
Another streetcar passed by. There was only a small space on the platform, making it the most suitable one to board early and hide in the middle. I clicked my tongue and lit a cigarette. The streetcar briefly stopped in front of the station where ○'s wife usually boarded and then departed again. But it stopped again after moving just one section. I absentmindedly threw away the \_ I had started consuming and turned toward that side. \\
\\
\textbf{Choices:} \\
- cigarette \\
- wife \\
\\
\textbf{Answer:} \\
cigarette
}
\\ \midrule

\textbf{Context1} & \tablecell{
전차는 또 한 대 지나갔다. 승강대에 빈틈이 조금 있을 뿐, 미리 올라타서 가운데 숨어 있기에는 가장 적절한 전차였었다. 나는 혀를 한번 차고 담배를 꺼내어 붙여 물었다. 그 전차는 ○의 아내의 타는 정류장 앞에서 잠깐 멎었다가 다시 떠났다. 그러나 한 간을 나아가지 않아서 그 전차는 다시 멎었다. 나는 무심히 먹기 시작한 \textbf{담배}
} & \tablecell{
Another streetcar passed by. There was only a small space on the platform, making it the most suitable one to board early and hide in the middle. I clicked my tongue and lit a cigarette. The streetcar briefly stopped in front of the station where ○'s wife usually boarded and then departed again. But it stopped again after moving just one section. I absentmindedly threw away the \textbf{cigarette}
}
\\ \midrule

\textbf{Context2} & \tablecell{
전차는 또 한 대 지나갔다. 승강대에 빈틈이 조금 있을 뿐, 미리 올라타서 가운데 숨어 있기에는 가장 적절한 전차였었다. 나는 혀를 한번 차고 담배를 꺼내어 붙여 물었다. 그 전차는 ○의 아내의 타는 정류장 앞에서 잠깐 멎었다가 다시 떠났다. 그러나 한 간을 나아가지 않아서 그 전차는 다시 멎었다. 나는 무심히 먹기 시작한 \textbf{아내}
} & \tablecell{
Another streetcar passed by. There was only a small space on the platform, making it the most suitable one to board early and hide in the middle. I clicked my tongue and lit a cigarette. The streetcar briefly stopped in front of the station where ○'s wife usually boarded and then departed again. But it stopped again after moving just one section. I absentmindedly threw away the \textbf{wife}
}
\\ \midrule

\textbf{Endings} & \tablecell{
를 내어던지고 그편을 향하여 돌아섰다.
} & \tablecell{
and turned toward that side.
}
\\ \bottomrule

\end{tabular}%
\caption{Evaluation prompts for SNU-Ko-LAMBADA}
\label{tab:eval-prompts-snu-lambada}
\end{table*}

\end{document}